%% file: main_ijcai21-multiauthor.tex
\documentclass{article}
\usepackage{ijcai21}

\usepackage{times}

\usepackage{soul}
\usepackage{url}
\usepackage[hidelinks]{hyperref}
\usepackage[utf8]{inputenc}
\usepackage[small]{caption}
\usepackage{graphicx}
\usepackage{amsmath}
\usepackage{booktabs}
\usepackage{kotex}
\usepackage{amsmath}
\usepackage{amsthm}
\usepackage{multirow}
\usepackage{array}
\usepackage{hhline}
\usepackage[flushleft]{threeparttable}
\usepackage[title]{appendix}
\usepackage{verbatim} 
\usepackage[linesnumbered,lined,ruled,noend]{algorithm2e}
\usepackage{subfigure}

\newcolumntype{?}{!{\vrule width 1pt}}

\DeclareMathOperator{\argmin}{argmin}

\newcommand{\eg}{\textit{e.g.}}
\newcommand{\ie}{\textit{i.e.}}

\title{Accelerating Neural Architecture Search via Proxy Data}

\author{
Byunggook Na$^1$\and
Jisoo Mok$^1$\and
Hyeokjun Choe$^1$\And
Sungroh Yoon$^{1,2}$\footnote{Contact Author}\\

\affiliations
$^1$ Department of Electrical and Computer Engineering, Seoul National University, Seoul, South Korea\\
$^2$ AIIS, ASRI, INMC, and Interdisciplinary Program in AI, Seoul National University, Seoul, South Korea \\

\emails
\{byunggook.na, jmok908, hyeokjun.choe\}@gmail.com,
sryoon@snu.ac.kr
}

\begin{document}

\maketitle

\begin{abstract}
\input{tex/0_abstract}

\end{abstract}

\section{Introduction}
\input{tex/1_intro}

\section{Related Work}
\input{tex/2_related_works}
\section{Exploration Study}\label{sec:exploration}
\input{tex/3_exploration_study}
\section{Proposed Selection Method}\label{sec:proposed}
\input{tex/4_proposed_method}

\section{Experiments and Results}\label{sec:results}
\input{tex/5_results}

\section{Conclusion}
\input{tex/6_conclusion}

\section*{Acknowledgements}
This work was supported by the National Research Foundation of Korea (NRF) grant funded by the Korea government (MSIT) [2018R1A2B3001628], Samsung Electronics Co., Ltd, and the Brain Korea 21 Plus Project in 2021.

\clearpage
\bibliographystyle{named}
\bibliography{references_abbr}

\clearpage
\appendix
\setcounter{figure}{0}
\setcounter{equation}{0}
\setcounter{table}{0}
\renewcommand{\thefigure}{A\arabic{figure}}
\renewcommand{\theequation}{A\arabic{equation}}
\renewcommand{\thetable}{A\arabic{table}}

\input{tex/8_appendix}


\end{document}

%% file: tex/0_abstract.tex

Despite the increasing interest in neural architecture search (NAS), the significant computational cost of NAS is a hindrance to researchers.  
Hence, we propose to reduce the cost of NAS using proxy data, \ie, a representative subset of the target data, without sacrificing search performance.
Even though data selection has been used across various fields, our evaluation of existing selection methods for NAS algorithms offered by NAS-Bench-1shot1 reveals that they are not always appropriate for NAS and a new selection method is necessary.
By analyzing proxy data constructed using various selection methods through data entropy, we propose a novel proxy data selection method tailored for NAS.
To empirically demonstrate the effectiveness, we conduct thorough experiments across diverse datasets, search spaces, and NAS algorithms.
Consequently, NAS algorithms with the proposed selection discover architectures that are competitive with those obtained using the entire dataset.
It significantly reduces the search cost: executing DARTS with the proposed selection requires only 40 minutes on CIFAR-10 and 7.5 hours on ImageNet with a single GPU.
Additionally, when the architecture searched on ImageNet using the proposed selection is inversely transferred to CIFAR-10, a state-of-the-art test error of 2.4\% is yielded.
Our code is available at https://github.com/nabk89/NAS-with-Proxy-data.

%% file: tex/1_intro.tex
Neural architecture search (NAS), one of the most widely-studied fields in automated machine learning, aims to reduce the human cost of designing and testing hundreds of neural architectures. 
In early studies pertaining to NAS~\cite{le2017naswithRL,zoph2018nasnet,real2019regularized}, however, enormous computational overhead occurred, thereby requiring a significant amount of computing resources to execute search algorithms.
To reduce the search cost, most recently developed NAS algorithms employ weight-sharing on a super-network and/or differentiable approach to optimize the architecture parameters in the super-network~\cite{xie2020survey}. 
These techniques enable a more approximate yet faster evaluation of candidate neural architecture performance, thereby significantly reducing the cost of NAS.


In this study, we further reduce the search cost of NAS by incorporating proxy data, \ie, a representative subset of the target data.
Data selection is widely used across various fields in deep learning, such as active learning~\cite{settles.tr09activesurvey,coleman2019selection} and curriculum learning~\cite{graves2017autocurriculum,chang2017activebias}.
For instance, given a trained model, the core-set selection methods used in active learning aim to select the training data from a large unlabeled dataset to label the selected data with minimum labeling costs, resulting in the effective reduction of the computational cost.
However, comprehensive studies regarding the problem of data selection for NAS do not exist.
Developing an appropriate selection for NAS is important because NAS algorithms could benefit from the significant reduction in the search cost.
We first evaluate the proxy data constructed using five existing data selection methods on NAS-Bench-1shot1~\cite{zela2020bench1shot1}.
While the substantial results provide strong empirical support for our hypothesis, they also reveal the necessity for a new, improved selection method, designed specifically for NAS.
Subsequently, we analyze the relationship between the search performances and properties of different selection methods based on the entropy~\cite{shannon@entropy} of examples in the proxy data.
Based on our analysis, the characteristics of a selection method that renders proxy data effective in preserving the search performance of NAS algorithms are identified.
When the selection method chooses primarily low-entropy examples, a competitive architecture is discovered with the resulting proxy data, even when the size of the proxy data is small.
To achieve the search performance obtained using the entire data, it is important to include additional high-entropy examples in the proxy data.

Based on these observations, we propose a new selection method that prefers examples in the tail ends of the data entropy distribution.
The selection method can be implemented in a deterministic or probabilistic manner.
For the former, we adopt the ratio between low- and high-entropy examples, such that the examples from the opposite ends of the entropy distribution are selected.
For the probabilistic manner, we suggest three sampling probabilities that satisfy the identified characteristics of the proxy data effective for NAS.
We demonstrate the superiority of the proposed selection to existing selections using NAS-Bench-1shot1 and show that the search performance is preserved even when using only 1.5K of training examples selected from CIFAR-10.

We further demonstrate that the proposed selection method can be applied universally across various differentiable NAS algorithms and four benchmark datasets for image classification.
The NAS algorithms with the proposed selection discover competitive neural architectures in a significantly reduced search time. For example, executing DARTS~\cite{yang2019darts} using our selection method requires only \textbf{40 GPU minutes} on a single GeForce RTX 2080ti GPU.
Owing to the reduction in search cost, searching on ImageNet can be completed in \textbf{7.5 GPU hours} on a single Tesla V100 GPU when incorporating the proposed selection into DARTS. 
The searched architecture achieves the top-1 test error of 24.6\%, surpassing the performance of the architecture, which is discovered by DARTS on CIFAR-10 and then transferred to ImageNet. 
In addition, when this architecture is evaluated on CIFAR-10, it achieves a top-1 test error of \textbf{2.4\%}, demonstrating state-of-the-art performance on CIFAR-10 among recent NAS algorithms. 
This indicates that the architectures discovered by NAS algorithms with proxy data selected from a large-scale dataset are highly transferable to smaller datasets.

To summarize, our main contributions are as follows:
\begin{itemize}
    \item We provide substantial experimental results conducted on NAS-Bench-1shot1 to demonstrate that the existing selection is inappropriate for NAS. 
    \item By identifying the characteristics of effective proxy data selection methods, we propose a novel selection method and two approaches for implementing it.
    \item We demonstrate the efficacy of the proposed selection for NAS and its general applicability to various NAS algorithms and datasets. We expect the field of NAS to benefit significantly from the reduced search cost afforded using the proposed proxy data selection.
\end{itemize}

%% file: tex/2_related_works.tex
\subsection{Neural Architecture Search}
Recently, NAS methods have become diversified significantly, as more complex algorithms have been developed to achieve higher performance or lower search cost~\cite{xie2020survey}. 
Herein, we discuss studies that focus on reducing the search cost.
To reduce the search cost, most differentiable and one-shot NAS methods have adopted weight-sharing~\cite{dean2018enas} or a continuous architecture with mixing weights~\cite{yang2019darts} on a cell-based search space~\cite{zoph2018nasnet}.
During the search, a super-network, which stacks multiple cells but is smaller than the target network, is trained.


To further reduce the cost, PC-DARTS~\cite{xu2020pcdarts} reduced the number of trainable weight parameters in the cells used during the search by partially bypassing channels in a shortcut.
EcoNAS~\cite{zhou2020econas} suggested four reduction factors, resulting in a much smaller network than the super-network of conventional cell-based NAS algorithms, and proposed a hierarchical evolutionary algorithm-based search strategy to improve the accuracy of architecture performance estimation using the smaller super-networks.
Regarding data selection, EcoNAS briefly reported the search result using a subset randomly sampled from CIFAR-10. 
In this study, we evaluate various selection methods, including random selection, and provide meaningful insights into the selection method tailored to NAS.


NAS algorithms suffer from the lack of reproducibility, and hence, a fair comparison of NAS algorithms is challenging.
Benchmarks for NAS~\cite{ying2019bench101,zela2020bench1shot1,dong2020nasbench201,dong2020nats} aim to alleviate this issue in NAS research. 
Because these benchmarks provide architecture databases and easy-to-implement NAS algorithm platforms, re-training searched architectures for evaluation can be omitted.
Therefore, in this study, we utilize two benchmarks, \ie, NAS-Bench-1shot1~\cite{zela2020bench1shot1} and NATS-Bench~\cite{dong2020nats}, to investigate selection methods for constructing effective proxy data and evaluate our proxy data selection method.


\subsection{Data Selection in Deep Learning}
Data selection or subsampling is a well-established methodology in deep learning, and it is used across various fields in deep learning.
Active learning~\cite{settles.tr09activesurvey,sener2018active,beluch2018power,coleman2019selection} adopts core-set selections to reduce the labeling cost by selecting the smallest possible number of examples from a large unlabeled dataset.
As an application, the approach in core-set selections can be applied to reduce the batch size for training generative adversarial networks~\cite{sinha2019small}.
In curriculum learning~\cite{graves2017autocurriculum,chang2017activebias}, examples are fed into a neural network efficiently to avoid catastrophic forgetting and accelerate training; hence, curriculum learning ends up utilizing the entire dataset, rather than a subset.
However, our results reveal that existing selection methods are not always appropriate for NAS; as such, a new selection method specifically for NAS is necessary.

%% file: tex/3_exploration_study.tex
\begin{figure*}[t]
    \centering
    \includegraphics[width=\linewidth]{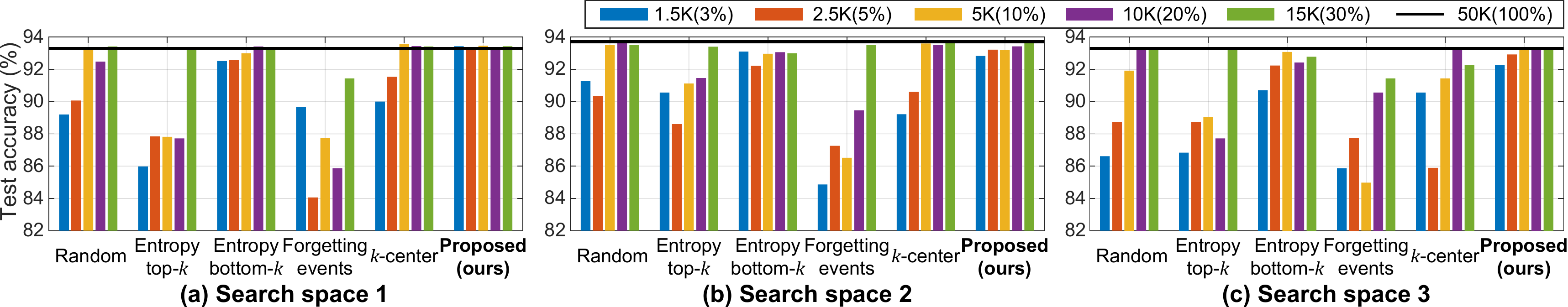}
    \vspace{-1.5em}
    \caption{Search performance (CIFAR-10 test accuracy) on NAS-Bench-1shot1 (DARTS) with various proxy data. \textbf{Proposed (ours)} indicates the proposed selection method in Section~\ref{sec:proposed}, specifically, the probabilistic selection method based on $P_{1}$.}
    \label{fig:nas-bench-1shot1_darts}
    \vspace{-0.7em}
\end{figure*}
\begin{figure*}[t]
    \centering
    \includegraphics[width=\linewidth]{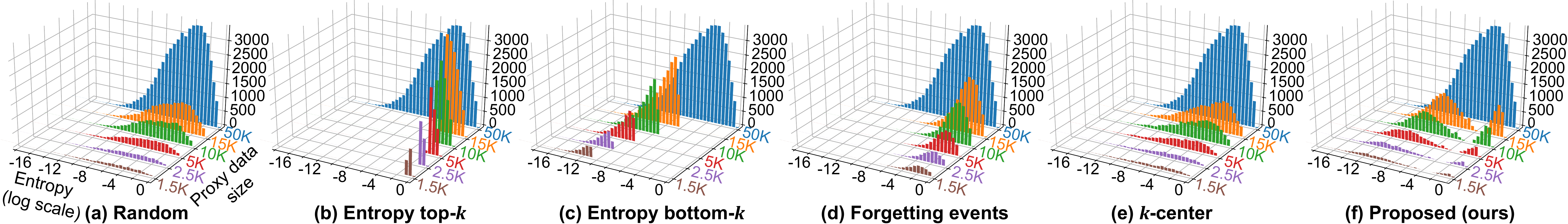}
    \vspace{-1.5em}
    \caption{Histograms of data entropy in log scale. Blue histograms indicate the entropy distributions of 50K training examples of CIFAR-10 and the others indicate those of various proxy data constructed using selection methods.}
    \label{fig:3d_histogram}
    \vspace{-1em}
\end{figure*}
In this study, we extensively evaluate the search performance of NAS algorithms offered by NAS-Bench-1shot1 using different proxy data constructed using existing selection methods.
Based on the results obtained, we identify the characteristics of selection methods that yield particularly effective proxy data for NAS.


\subsection{Exploration Setting}
NAS-Bench-1shot1 is used as the primary testing platform to observe the effect of the proxy data on the search performance of three NAS algorithms on CIFAR-10: DARTS~\cite{yang2019darts}, ENAS~\cite{dean2018enas}, and GDAS~\cite{dong2019searching}.
To construct proxy data of size $k$, $k$ examples among 50K training examples of CIFAR-10 are selected using selection methods.
The selected examples are segregated into two parts: one for updating weight parameters and the other for updating architecture parameters.
For a fair comparison, the same hyperparameter settings as those offered in NAS-Bench-1shot1 are used for all the tested NAS algorithms.
To avoid cherry-picking results, we execute the search process five times with different seeds and report the average performance.

The five selection methods utilized in this study are: random, entropy top-$k$, entropy bottom-$k$, forgetting events, and $k$-center.
Herein, we provide a brief description of each selection method; more details are included in the Appendix.
Random selection, as the name suggests, samples examples uniformly. 
For entropy-based selection~\cite{settles.tr09activesurvey}, we use the entropy value $f_\mathrm{entropy}$ of example $x$ calculated as: 
\begin{equation}\label{eq:entropy}
    f_\mathrm{entropy}(x; M) = - \sum\nolimits_{\boldsymbol{\Tilde{y}}} P(\Tilde{y}|x;M) \log{P(\Tilde{y}|x;M)},
\end{equation}
where $\boldsymbol{\Tilde{y}}=M(x)$ is the predictive distribution of $x$ with respect to the pre-trained baseline model $M$, \ie, the input of softmax in a classifier.
Entropy top-$k$ selection selects examples that have top-$k$ entropy, and entropy bottom-$k$ selection performs the opposite. 
For the forgetting event selection~\cite{toneva2018forgetting}, we train a model from scratch and monitor the number of misclassifications referred to as forgetting events per example. 
After training the model, examples whose number of forgetting events is in the top-$k$ are selected. 
In the $k$-center selection~\cite{sener2018active}, given feature vectors extracted from a pre-trained model for all examples, $k$ examples are selected by a greedy $k$-center algorithm.
We set $k \in \{1.5, 2.5, 5, 10, 15\}$K, and pre-train ResNet-20~\cite{he2016resnet} for the selection methods; the test accuracy of the pre-trained model is 91.7\%.
Proxy data constructed using the five selections are denoted by $D_\mathrm{random}$, $D_\mathrm{top}$, $D_\mathrm{bottom}$, $D_\mathrm{forget}$, and $D_\mathrm{center}$, correspondingly.

\subsection{Observations and Analysis}

The search results of DARTS are shown in \figurename~\ref{fig:nas-bench-1shot1_darts}, and those of ENAS and GDAS are provided in the Appendix.
As $k$ changes, two interesting phenomena are observed.
First, for $k \leq 2.5$K examples, searching with $D_\mathrm{bottom}$ consistently yields a search performance closer to that of DARTS with the entire data, namely the original performance.
Second, as $k$ increases above $5$K, searching with most proxy data results in a rapid increase in the resulting search performance; in the case of $D_\mathrm{bottom}$, however, the improvement is less prominent, and the original performance is hardly achieved. 

We analyze different proxy data to identify the most significant factor that contributes to the search performance competitive to the original performance using data entropy, $f_\mathrm{entropy}$.
Data entropy is a typically used metric to quantify the difficulty of an example; furthermore, it is used as the defining property of proxy data in this study.
\figurename~\ref{fig:3d_histogram} shows the histograms of data entropy of all proxy data in log scale.



As shown in \figurename~\ref{fig:3d_histogram}(a)-(e), the composition of $D_\mathrm{bottom}$ differs significantly from those of the other proxy data.
When $k \leq 2.5$K, $D_\mathrm{bottom}$, which achieves a more competitive search performance than other proxy data, contains a significantly larger number of easy examples.
It suggests that to construct proxy data with a number of easy examples is effective for minimizing the size of proxy data and obtaining the original search performance.
Meanwhile, $D_\mathrm{random}$ (or $D_\mathrm{center}$) gradually includes easy, middle, and difficult examples, and most additional examples in $D_\mathrm{forget}$ (or $D_\mathrm{top}$) are difficult.
It appears that NAS with the proxy data, which appropriately includes middle and difficult examples, can achieve the original search performance when $k$ is sufficiently large; the appropriate value of $k$ differs for each selection.
Comprehensively, based on the observed correlations in the search performance and the compositions in the proxy data, we deduce that selection methods for NAS satisfy the following characteristics:
    
    $\bullet$ When a small number of examples are selected, easy examples are more likely to discover a relatively competitive architecture than difficult examples.
    
    $\bullet$ When easy examples are already selected, adding middle and difficult examples enables the original search performance to be achieved.

%% file: tex/4_proposed_method.tex
\begin{figure}[t]
    \centering
    \includegraphics[width=\linewidth]{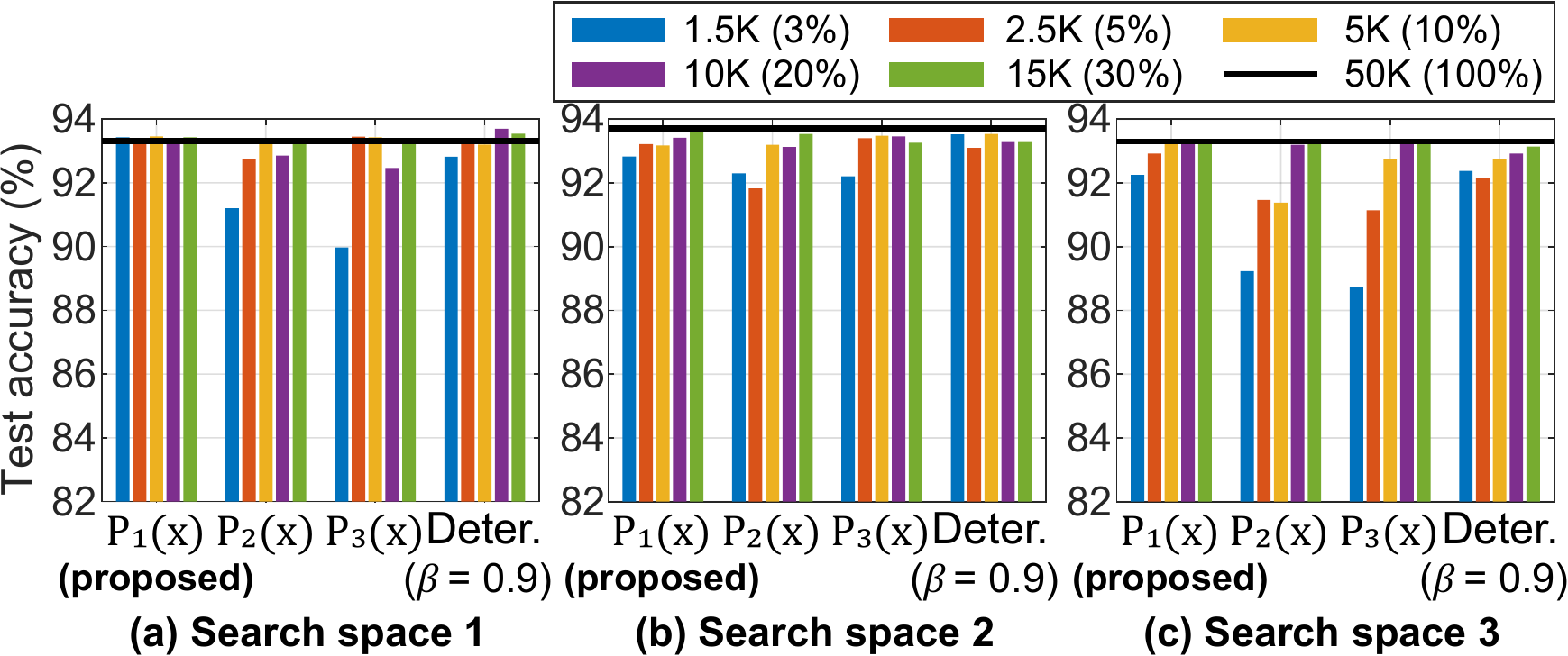}
    \caption{Search performance on NAS-Bench-1shot1 (DARTS) using the proposed methods. $P_{\{1,2,3\}}(x)$ are sampling probabilities used in the proposed probabilistic selection, and Deter. indicates the proposed deterministic selection with $\beta=0.9$. }
    \label{fig:darts_proposed}
    \vspace{-1em}
\end{figure}

\figurename~\ref{fig:nas-bench-1shot1_darts} shows that the size of the smallest effective proxy data obtained using the existing selection methods is $5$K.
Although random selection may be considered a strong baseline selection method, its performance deteriorates significantly when $k \leq 2.5$K.
In addition, it is noticeable that $D_\mathrm{bottom}$ with $k \leq 2.5$K achieves the better search performance than the other proxy data.
Therefore, to further minimize the size of the proxy data, we propose a new selection method that weighs on examples belonging to both-sided tailed distribution in the data entropy.
It is intuitive that the increase in difficult examples provides more information to NAS with easy examples than additional middle examples.
As shown in \figurename~\ref{fig:nas-bench-1shot1_darts}, this is supported by results of $D_{bottom}$ with $k=15$K, which includes a large number of easy examples and a small number of middle examples.


Herein, we suggest two methods for implementing the proposed selection method: deterministic and probabilistic methods.
For deterministic selection, we adopt the composition ratio parameter $\beta$ of low-entropy examples.
With respect to data entropy, bottom-$\beta k$ examples and top-$(1-\beta) k$ examples are selected, where $0<\beta<1$.
For probabilistic selection, the probability distribution of examples should be designed to satisfy the two aforementioned characteristics.
Utilizing histogram information, which can be obtained using a pre-trained baseline model, we design and evaluate three probabilities, denoted by $P_1$, $P_2$, and $P_3$, for probabilistic selection.
Let $h_x$ denote a bin where example $x$ belongs in data entropy histogram $H$, and $|h_x|$ denote the height of $h_x$, \ie, the number of examples in $h_x$.
The three probabilities are defined as follows:
\begin{equation}\label{eq:prob}
    P_{\{1,2,3\}}(x; H) = \mathrm{norm(} W_{\{1,2,3\}}(h_x; H) / |h_x| \mathrm{)},
\end{equation}
where $\mathrm{norm()}$ normalizes the inside term such that $\sum_{x \in D} P_{\{1,2,3\}}(x;H) = 1$ for target data $D$.
In the inside term, selection weights denoted by $W_{\{1,2,3\}}(h_x; H)$ are defined as follows:
\begin{equation}
    W_{1}(h_x; H) = \frac{\mathrm{max}_{h'\in H}{|h'|} - |h_x| + 1} {\sum_{h'' \in H} \mathrm{max}_{h'\in H}{|h'|} - |h''| + 1} ,
\end{equation}
\begin{equation}
    W_{2}(h_x; H) = \frac{1} {\text{the number of bins in } H} ,
\end{equation}
\begin{equation}
    W_{3}(h_x; H) = \frac{1/|h_x|} {\sum_{h'' \in H} 1/|h''|} .
\end{equation}
In Eq.~\ref{eq:prob} with $W_{2}(h_x; H)$, which places equal weights on all bins, examples from tail-ends of $H$ are more likely to be selected. 
$W_{1}$ and $W_{3}$ further penalize middle examples by using the difference between height of $h_x$ and the maximum height of the bin near the center in $H$.


For evaluation on NAS-Bench-1shot1, we execute the proposed selections using 10 different seeds.
Among the deterministic selections with $\beta = \{0.9, 0.8, 0.7, 0.6, 0.5\}$, the search performance with $\beta=0.9$ is the best; the other results are provided in the Appendix.
For probabilistic selection, we quantify $|h_x|$ based on a data entropy histogram of CIFAR-10, which is the blue histogram in \figurename~\ref{fig:3d_histogram}.
\figurename~\ref{fig:3d_histogram}(f) shows that the entropy distribution of examples selected by the proposed selection. 

As shown in \figurename~\ref{fig:darts_proposed}, among the selections, the deterministic selection with $\beta=0.9$ and the probabilistic selection based on $P_1(x;H)$ achieve the best search performance.
In particular, in search space 1, the $P_1(x;H)$-based probabilistic selection achieves the original performance with only $k=1.5$K examples. 
Although the deterministic selection with $\beta=0.9$ achieves a competitive performance as well, finding the optimal $\beta$ is nontrivial because the optimal $\beta$ can be dependent on the target data or pre-trained baseline models. 
By contrast, probabilistic selection does not require additional hyperparameters; as such, an exhaustive hyperparameter search is not necessary for selecting proxy data.
Therefore, we set the $P_1(x;H)$-based probabilistic selection as our main method for the remainder of the study.

As shown in \figurename~\ref{fig:nas-bench-1shot1_darts}, the proposed proxy data selection demonstrates better search performance compared with the other existing selections.
We include evaluation results on NATS-Bench~\cite{dong2020nats} in the Appendix, where the results also show our method is valid on NAS algorithms offered by NATS-Bench.
We further demonstrate its effectiveness on an enlarged cell-based search space~\cite{yang2019darts} and various NAS algorithms in Section~\ref{sec:results}.

\subsection{Discussion Regarding Efficacy of Proposed Selection}

In this section, the factor contributing to the effectiveness of the proposed selection method particularly for NAS is discussed.
Many differentible NAS algorithms focus on training a super-network~\cite{xie2020survey}.
When a super-network is trained to fit only the easier examples, it will naturally converge faster than when it is attempting to fit difficult examples. 
The side effect of this phenomenon is that the gradient of the loss will become small only after a few epochs of training~\cite{chang2017activebias} and hence will no longer backpropagate useful signals for the super-network.
Therefore, when deriving an architecture from such super-network,
it is likely that the resulting architecture will have limited generalization capacity to difficult examples. 
Using difficult examples allows the super-network to learn more meaningful information, which is difficult to be obtained from easy examples.
Using the t-SNE~\cite{maaten2008tsne} visualization of different proxy data, we can speculate that the missing information from the easy examples is related to the decision boundaries obtained from the pre-trained network and the dataset.
As mentioned in Section~\ref{sec:proposed}, we use ResNet-20 to extract features from different proxy data of CIFAR-10; the corresponding t-SNE visualization results are shown in the Appendix. 
The easy examples tend to be distant from the decision boundaries, unlike the difficult ones.
Therefore, for a super-network to learn such decision boundaries, proxy data with difficult examples is required.

Meanwhile, if proxy data is comprised primarily of difficult examples, the stable training of a super-network may be hindered, which is consistent with the results of $D_{top}$ in \figurename~\ref{fig:nas-bench-1shot1_darts}.
This issue can be resolved using a sufficient number of easy examples.
Consequently, the proposed selection method that satisfies the characteristics identified in Section~\ref{sec:exploration} yields a super-network that is similar to that trained with the entire dataset while the size of the proxy data is minimized.

%% file: tex/5_results.tex
\begin{figure}[t]
    \centering
    \includegraphics[width=\linewidth]{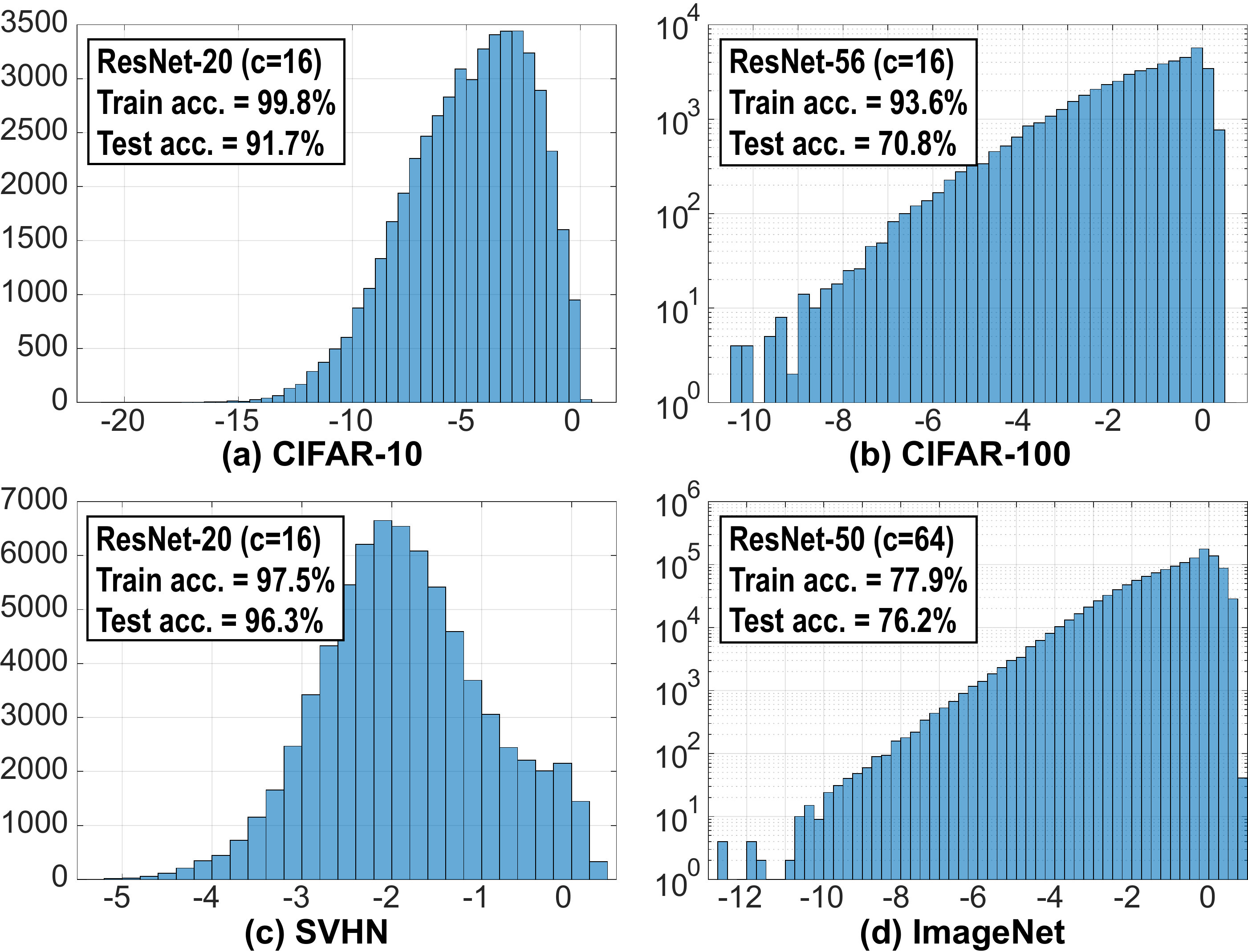}
    \vspace{-1em}
    \caption{Histograms of data entropy (log scale in x-axis). In each histogram, \textbf{c} indicates the number of filters in the first convolutional layer in each ResNet. The histograms from (a) to (d) are obtained by setting a bin width as 0.5, 0.25, 0.2, and 0.25, respectively.}
    \label{fig:histogram}
    \vspace{-1em}
\end{figure}


We used the pre-trained ResNet-50 in Pytorch model zoo for ImageNet, and trained the three models for CIFAR-10, CIFAR-100, and SVHN; the training took 0.015, 0.033, and 0.022 GPUdays, where the additional cost is negligible.
We prepared proxy data using log-scaled data entropy histograms in \figurename~\ref{fig:histogram}.
Although the data entropy distributions of CIFAR-100 and ImageNet show different patterns than those of CIFAR-10 and SVHN, our experimental results consistently indicate that the proposed proxy data selection is valid for CIFAR-100 and ImageNet, as it is for the other two datasets.
For the evaluation, we used two types of GPUs: Tesla V100 for searching and training neural architectures on ImageNet, and GeForce RTX 2080ti for the remaining datasets.
We execute the search processes using three different seeds and report the averaged values.
More details regarding the experimental settings are included in the Appendix.


\subsection{Comparison with Random Selection}
\input{tex/5_results_random_vs_proposed}
Based on Section~\ref{sec:exploration}, it is apparent that random selection is an effective, reasonable baseline.
Hence, we compare the proposed selection with random selection in the cell-based search space using DARTS~\cite{yang2019darts} with CIFAR-10; search results with the other selections evaluated in Section~\ref{sec:exploration} are included in the Appendix.
We change the size of proxy data from 5K to 25K; the search cost decreases proportionally to the size of proxy data.
As shown in \tablename~\ref{table:random_vs_proposed}, on CIFAR-10, searching with the proposed method is usually superior to that using random selection.
Furthermore, the search performance with the random selection fluctuates with varying sizes of proxy data.
Result of searching with 128K training examples chosen from ImageNet by the proposed selection, is also superior to that of random selection.

\subsection{Applicability to NAS Algorithms}
\input{tex/5_results_real_nas}
Recently, various differentiable NAS algorithms based on a cell-based search space have been proposed~\cite{xie2020survey}.
We apply the proposed proxy data selection to the recently proposed NAS, \ie, DARTS~\cite{yang2019darts}, PC-DARTS~\cite{xu2020pcdarts}, SDARTS~\cite{chen2020stabilizing-darts}, SGAS~\cite{li2020sgas}, and EcoDARTS that is a DARTS-based variant of EcoNAS~\cite{zhou2020econas}, respectively.
As shown in \tablename~\ref{table:cell_based_real_nas}, all of the tested NAS algorithms achieve the comparable performance to their respective original search performance.
While on CIFAR-10, PC-DARTS with the propose selection experiences a slight decrease in performance, on ImageNet, it succesfully achieves the original search performance with significantly reduced search cost.

None of the existing NAS algorithms searched directly on ImageNet, with the exception of PC-DARTS.
To perform the direct search on ImageNet, we incorporate the proposed selection with PC-DARTS and DARTS. 
DARTS with the proposed selection discovers a better architecture than the original DARTS, which transfers the architecture searched on CIFAR-10 to ImageNet. 
The original PC-DARTS searched on ImageNet with 12.5\% of examples randomly sampled from the dataset.
It required 3.8 GPU days, \ie, 11.5 hours with eight V100 GPUs, with a batch size of 1024; we speculate that the parallel execution on the eight GPUs resulted in a non-negligible overhead.
By contrast, PC-DARTS with the proxy data which consists of 10\% of examples constructed using the proposed selection, can discover the competitive architecture using a single V100 GPU with a batch size of 256 in approximately 0.26 GPU days, \ie, 14.6 times less cost than that of the original PC-DARTS.

\subsection{Applicability to Datasets}
\input{tex/5_results_robustdarts}
%
To demonstrate the general applicability of the proposed selection to datasets, we test it on CIFAR-10, CIFAR-100, and SVHN using DARTS and RobustDARTS~\cite{Zela2020rdarts} in four different search spaces. 
These search spaces were modified from the cell-based search space by reducing the types of candidate operations (S1-S3) and inserting harmful noise operation (S4); the details are provided in the Appendix.
Following the experimental protocols in RobustDARTS, the weight decay factors for DARTS and RobustDARTS (L2) during search are set to be 0.0003 and 0.0243, respectively.

As shown in \tablename~\ref{table:robust_darts}, most results of the two NAS algorithms using the proposed selection are within a reasonable range of the original search performance.
However, when DARTS is executed on S4 with 10\% of examples from CIFAR-100, a significant search failure occurs.
This failure is caused because noise operations in S4 occupy most of the edges in the cell structure after search.
Note that the noise operation is intended for inducing failure in DARTS~\cite{Zela2020rdarts} and is generally not used in practice.
Nevertheless, the original search performance on CIFAR-100 can be obtained when using 20\% of examples.
\subsection{Inverse Transferability}
Typically, in most NAS algorithms, the transferability of architectures discovered using CIFAR-10 is demonstrated by their performance on ImageNet.
Using DARTS with the proposed selection, the computational cost of searching with ImageNet is reduced by $\frac{1}{10}$, \ie, 0.26 GPU days.
The resulting search time on the proxy data of ImageNet is similar to those of other NAS algorithms on the entire CIFAR-10.
Therefore, granted the similar amounts of search cost for fair comparison, architectures discovered on ImageNet using the proposed selection can be evaluated on CIFAR-10, which is the inverse way from conventional studies.

Consequently, the architecture searched on ImageNet using the proposed selection yields a top-1 test error of \textbf{2.4\%} on CIFAR-10, \ie, the best performance among cell-based NAS algorithms; the results with recent NAS algorithms are provided in the Appendix.
It is noteworthy that we do not utilize additional techniques introduced in recent studies, and that the architecture above is discovered only by executing DARTS on the proxy data of a large-scale dataset.
We speculate that the use of ImageNet provides DARTS with more helpful visual representations than CIFAR-10.
We refer to such an approach of transferring an architecture from a large-scale dataset to a smaller dataset as \textit{inverse transfer}.
Our study reveals that if the search cost on a large-scale dataset is reasonably low, then the inverse transfer of an architecture can provide new directions for NAS research.

%% file: tex/5_results_random_vs_proposed.tex
\begin{table}[t]
    \centering
    \renewcommand{\arraystretch}{1}
        \begin{tabular}{c|ccccc|c}
            \toprule
            \multirow{2}{*}{\textbf{Selection}} & \multicolumn{5}{c|}{\textbf{CIFAR-10}} & \textbf{IN} \\
             & \textbf{5K} & \textbf{10K} & \textbf{15K} & \textbf{20K} & \textbf{25K} & \textbf{128K}\\
            \midrule
            Random & 3.21 & 2.95 & 2.99 & 2.72 & 3.22 & 25.2 \\
            Proposed & 2.94 & 2.92 & 2.88 & 2.78 & 2.76 & 24.6 \\
            \bottomrule
        \end{tabular}
    \vspace{-0.5em}
    \caption{Evaluation (top-1 test error (\%)) of DARTS with varying sizes of proxy data of CIFAR-10 and ImageNet denoted by \textbf{IN}.}
    \label{table:random_vs_proposed}
    \vspace{-0.7em}
\end{table}

%% file: tex/5_results_real_nas.tex
\begin{table}[t]
    \centering
    \renewcommand{\arraystretch}{1}
        \begin{tabular}{c|cc|cc}
            \toprule
            \textbf{NAS} & \multicolumn{2}{c|}{\textbf{Base}} & \multicolumn{2}{c}{\textbf{Proposed}} \\
            \textbf{algorithm} & \textbf{Err. (\%)} & \textbf{Cost} & \textbf{Err. (\%) } & \textbf{Cost} \\ 
            \midrule
            \multicolumn{5}{c}{\textbf{CIFAR-10 (Base: 50K, Proposed: 5K)}} \\
            \midrule
            DARTS & 3.00 & 0.26 & 2.94 & 0.03 \\
            PC-DARTS & 2.67 & 0.08 & 2.91 & 0.01 \\
            EcoDARTS-c4r2 & 2.80 & 0.23 & 2.81 & 0.02 \\
            SDARTS-RS & 2.67 & 0.23 & 2.83 & 0.03 \\
            SGAS-Cri.1 & 2.66 & 0.19 & 2.72 & 0.02 \\
            \midrule
            \multicolumn{5}{c}{\textbf{ImageNet (Base: 1.28M, Proposed: 128K)}} \\
            \midrule
            DARTS & 26.7 & - & 24.6 & 0.32 \\
            PC-DARTS & 24.2 & 3.8$^\dagger$ & 24.3 & 0.26 \\
            \bottomrule
        \end{tabular}
    \vspace{-0.5em}
    \caption{Evaluation of various NAS methods used on cell-based search space using proposed proxy data selection. Search cost is GPU days and single 2080ti GPU and V100 GPU are used for searching on CIFAR-10 and ImageNet, respectively. $^\dagger$Authors of PC-DARTS reported that search process on ImageNet required 11.5 hours with eight V100 GPUs, \ie, 3.8 GPU days.}
    \label{table:cell_based_real_nas}
    \vspace{-0.7em}
\end{table}

%% file: tex/5_results_robustdarts.tex
\begin{table}[t]
    \centering
    \renewcommand{\arraystretch}{1.}
        \begin{tabular}{c|c|c|r|rr}
            \toprule
            \textbf{NAS} & \multirow{2}{*}{\textbf{Data}} & \textbf{Search} & \textbf{Base} & \multicolumn{2}{c}{\textbf{Proposed}} \\
            \textbf{alg.} & & \textbf{space} & \textbf{100\%} & \textbf{10\%} & \textbf{20\%} \\
            \midrule
            \multirow{12}{*}{\rotatebox[origin=c]{90}{DARTS}} & \multirow{4}{*}{CIFAR-10} & S1 & 3.84 & 3.60 & 2.96 \\
            &  & S2 & 4.85 & 3.54 & 3.46 \\
            &  & S3 & 3.34 & 2.71 & 2.72 \\
            &  & S4 & 7.20 & 6.60 & 5.82 \\
            \cmidrule{2-6}
            & \multirow{4}{*}{CIFAR-100} & S1 & 29.46 & 26.41 & 28.79 \\
            &  & S2 & 26.05 & 21.65 & 22.66 \\
            &  & S3 & 28.90 & 22.10 & 23.51 \\
            &  & S4 & 22.85 & 98.91 & 25.74 \\
            \cmidrule{2-6}
            & \multirow{4}{*}{SVHN} & S1 & 4.58 & 3.12 & 4.11 \\
            &  & S2 & 3.53 & 2.81 & 3.03 \\
            &  & S3 & 3.41 & 2.77 & 3.11  \\
            &  & S4 & 3.05 & 3.06 & 2.42 \\
            
            \midrule
            \multirow{12}{*}{\rotatebox[origin=c]{90}{RobustDARTS (L2)}} & \multirow{4}{*}{CIFAR-10} & S1 & 2.78 & 2.79 & 2.86 \\
            &  & S2 & 3.31 & 3.33 & 2.98 \\
            &  & S3 & 2.51 & 2.74 & 2.80 \\
            &  & S4 & 3.56 & 3.41 & 3.43 \\
            \cmidrule{2-6}
            & \multirow{4}{*}{CIFAR-100} & S1 & 24.25 & 26.13 & 23.67 \\
            &  & S2 & 22.24 & 22.21 & 21.39 \\
            &  & S3 & 23.99 & 21.71 & 22.31  \\
            &  & S4 & 21.94 & 27.83 & 21.10  \\
            \cmidrule{2-6}
            &\multirow{4}{*}{SVHN} & S1 & 4.79 & 2.46 & 2.60 \\
            &  & S2 & 2.51 & 2.45 & 2.49 \\
            &  & S3 & 2.48 & 2.53 & 2.42 \\
            &  & S4 & 2.50 & 5.16 & 2.62 \\
            \bottomrule
        \end{tabular}
    \vspace{-0.5em}
    \caption{Evaluation (top-1 test error (\%)) in four restricted cell-based search spaces and three datasets.}
    \label{table:robust_darts}
    \vspace{-1.1em}
\end{table}

%% file: tex/6_conclusion.tex
For the first time in NAS research, we introduced proxy data for accelerating NAS algorithms without sacrificing search performance.
After evaluating existing selection methods on NAS-Bench-1shot1, we obtained the insights and proposed a novel selection method for NAS, which prefers examples in tail-end of entropy distribution of the target data.
We thoroughly demonstrated the NAS acceleration and applicability of the proposed probabilistic selection on various datasets and NAS algorithms.
Notably, a direct search on ImageNet was completed in 7.5 GPU hours, suggesting that the inverse transfer approach is valid.
We expect other studies on NAS to benefit from the significant reduction in the search cost through the use of proxy data.



%% file: tex/8_appendix.tex
\section{Supplemental Materials of Selection Methods}

\subsection{Core-set Selection Methods}
\subsubsection{Random selection ($D_\mathrm{random}$)}
The first selection function randomly samples $k$ examples according to the uniform distribution.
By definition, $D_\mathrm{random}$ must inherit the characteristics of target data.

\subsubsection{Top-$k$ (bottom-$k$) entropy-based selection ($D_\mathrm{top}$, $D_\mathrm{bottom}$)}
These two selection functions~\cite{settles.tr09activesurvey} are based on the entropy of the predictive distribution of examples, abbreviated by the entropy of examples.
To construct $D_\mathrm{top}$ and $D_\mathrm{bottom}$, we calculate the entropy of all examples using a widely-known neural network (\eg, ResNet~\cite{he2016resnet}) pretrained on the target data.
Then, after ranking the examples according to their entropy in a descending order, we select the top-$k$ (bottom-$k$) examples. 
We calculate the entropy $f_\mathrm{entropy}(x)$ of example $x$ using a pretrained model $M$, as follows:
\begin{equation}
    f_\mathrm{entropy}(x; M) = - \sum_{\Tilde{y}} P(\Tilde{y}|x;M) \log{P(\Tilde{y}|x;M)},
\end{equation}
where $\boldsymbol{\Tilde{y}}=M(x)$ indicates the predictive distribution of $x$ with respect to $M$, $\boldsymbol{\Tilde{y}} \in R^d$, and $d$ is the output dimension of $M$.

\subsubsection{Forgetting events-based selection ($D_\mathrm{forget}$)} 
Forgetting events~\cite{toneva2018forgetting} of example $x$ indicate the number of times $x$ is misclassified after having been correctly classified during the training of a neural network.
Before constructing $D_\mathrm{forget}$, we train a widely-known neural network from scratch and keep track of the number of forgetting events per each example.
After training the neural network, the examples are ranked based on the number of forgetting events in descending order, and then top-$k$ examples are chosen from the ranked list.

\subsubsection{Greedy $k$-center selection ($D_\mathrm{center}$)} 
Intuitively speaking, the $k$-center problem aims to select $k$ center examples, such that the maximum distance between a given example and its nearest center is minimized~\cite{sener2018active}.
The formal definition of the $k$-center problem is given as:
\begin{equation}
    \min_{s^1:|s^1| \leq k} \max_{i} \min_{j \in s^1 \bigcup s^0} \Delta({x}_{i}, {x}_{j})
\end{equation}
where $s^{0}$ is the initial set of examples and $s^{1}$ is the newly selected set of examples.
To construct $D_\mathrm{center}$, we first extract features from each example by using a pretrained neural network.
Afterwards, given the features of examples, we select $k$ number of examples according to a greedy $k$-center selection approach described in \algorithmcfname~\ref{alg:kcenter}.

\begin{algorithm}[h]
\SetAlFnt{\small}
\SetAlCapFnt{\small}
\DontPrintSemicolon
\SetKwInOut{Input}{input}
\Input{target data $D$, initial random pool $P_{0} \subset D$, trained model $M$, and a budget $k$}
\BlankLine
	Initialize $P = P_{0}$, $D_\mathrm{center} = \emptyset$\;
	\textbf{repeat} \;
	~~~ $x = \mathrm{argmax}_{x_{i} \in D\backslash P}$ $\mathrm{min}_{x_{j} \in P}$ $\Delta (x_i, x_j; M)$ \;
	~~~ $P = P \cup \{x\}$ \;
	~~~ $D_\mathrm{center} = D_\mathrm{center} \cup \{x\}$ \;
	\textbf{until} $|D_\mathrm{center}| = k$\;
	\textbf{return} $D_\mathrm{center}$
\caption{Greedy $k$-center selection for $D_\mathrm{center}$ }
\label{alg:kcenter}
\end{algorithm}

\subsection{Data Entropy Histograms of Proxy Data}

\begin{figure*}[t]
    \centering
    \includegraphics[width=0.9\linewidth]{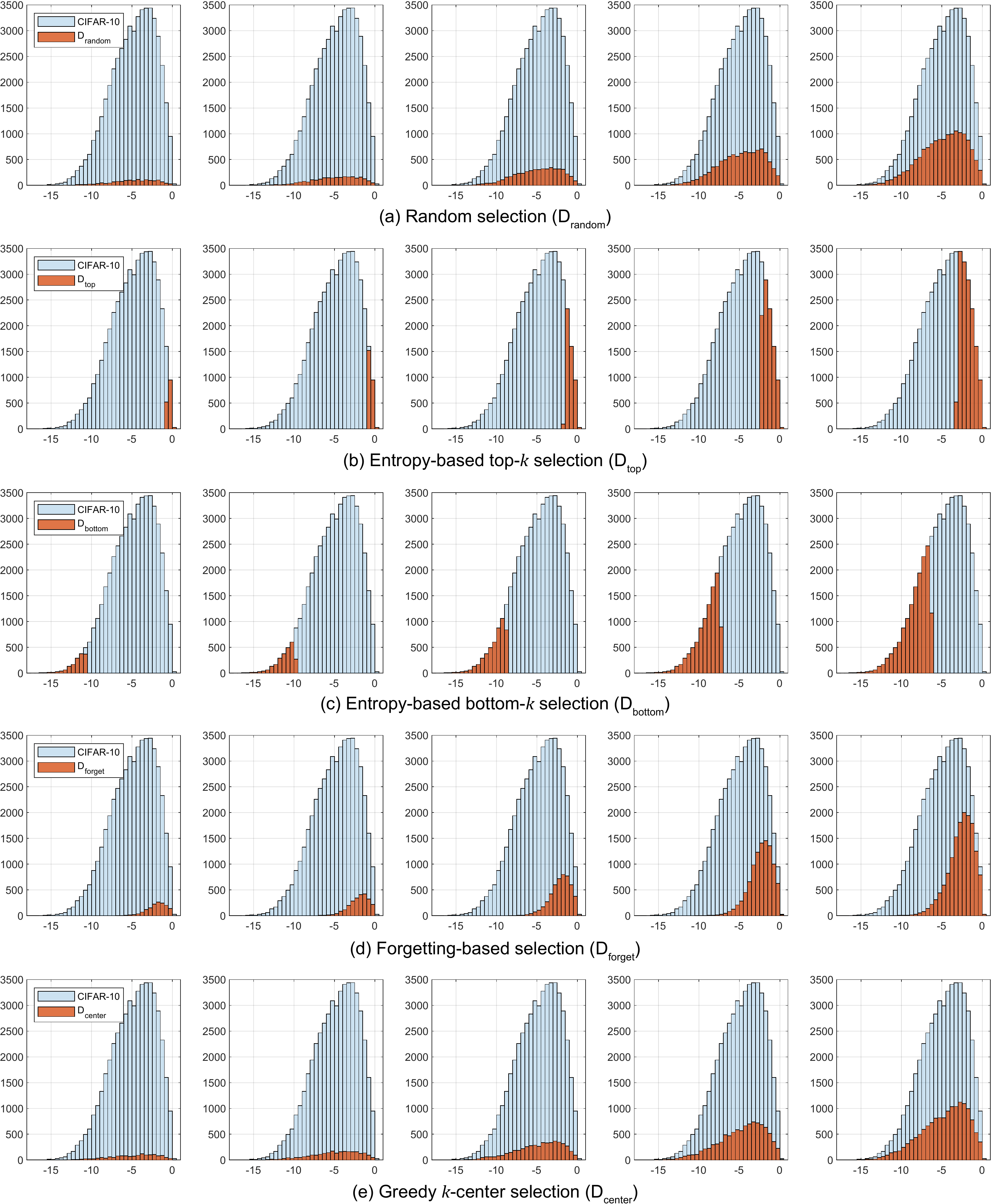}
    \caption{Entropy histograms of all proxy data we used to evaluate NAS algorithms of NAS-Bench-1shot1: (from left to right column) the number of examples in proxy data, \ie, $k = \{$1.5, 2.5, 5, 10, 15$\}$K.}
    \label{appendix_fig:ent_hist_proxy_merged}
    \vspace{-1.0em}
\end{figure*}

We attach \figurename~\ref{appendix_fig:ent_hist_proxy_merged} which shows individually data entropy histograms of all proxy data of CIFAR-10.

\subsection{Visualization of t-SNE}

\begin{figure*}[t]
    \centering
    \subfigure[]{
        \includegraphics[width=0.2\linewidth]{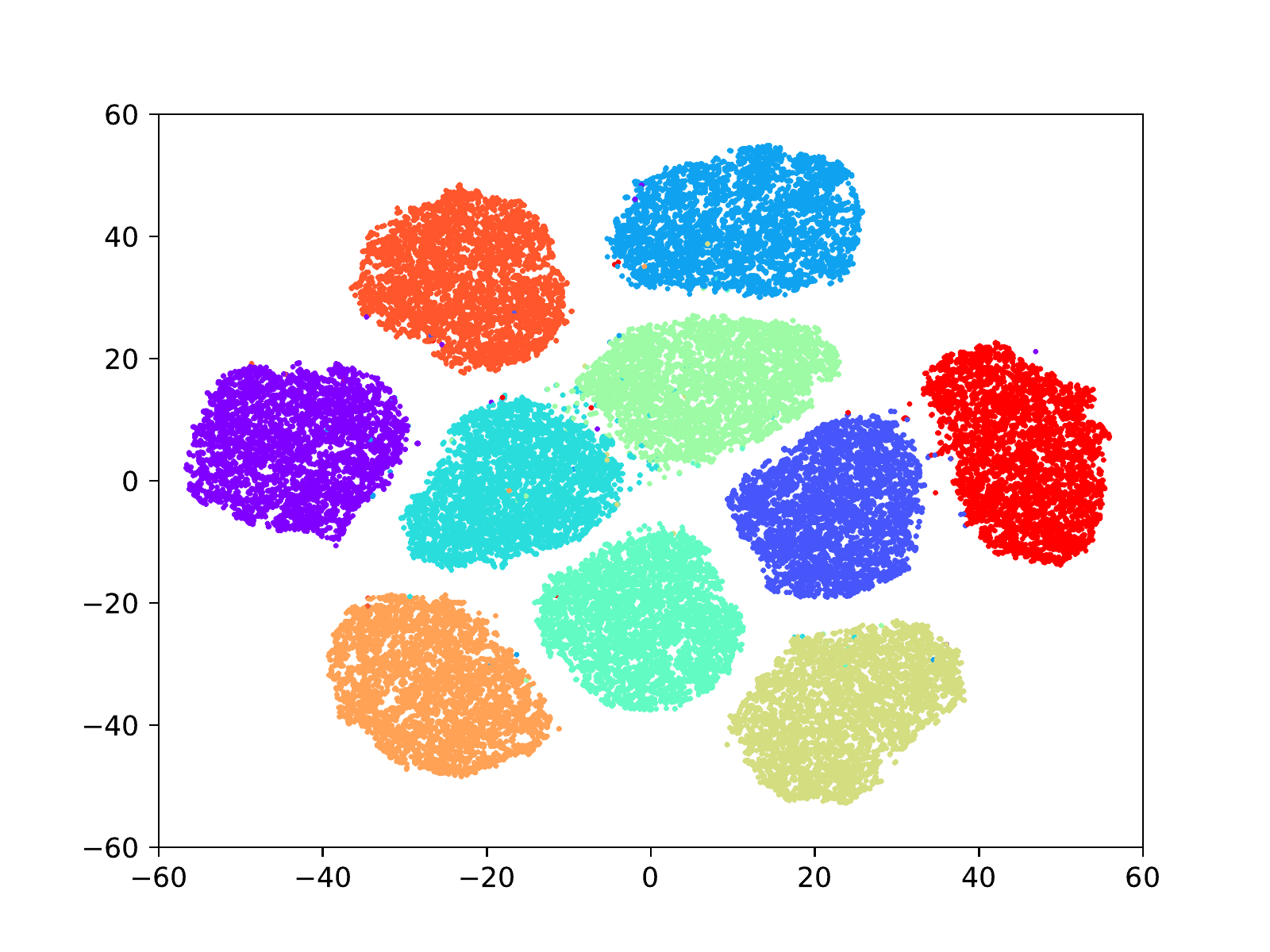}
    }
    \subfigure[]{
        \includegraphics[width=\linewidth]{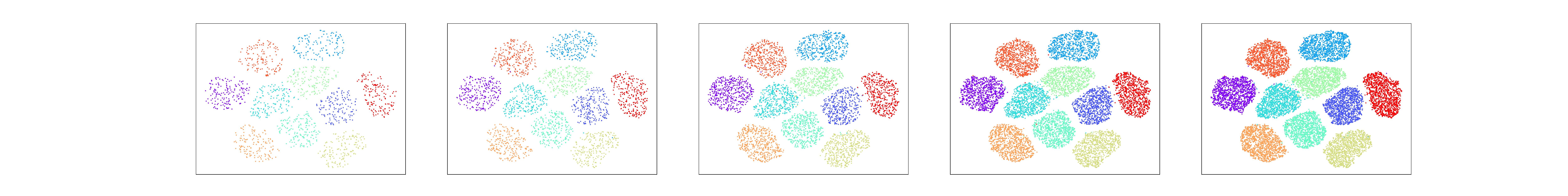}
    }
    \subfigure[]{
        \includegraphics[width=\linewidth]{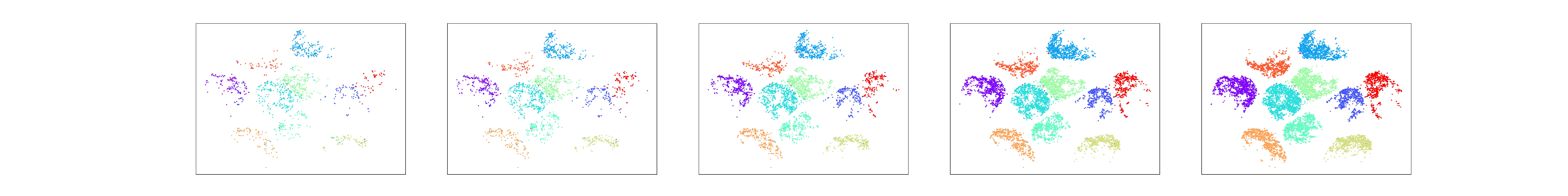}
    }
    \subfigure[]{
        \includegraphics[width=\linewidth]{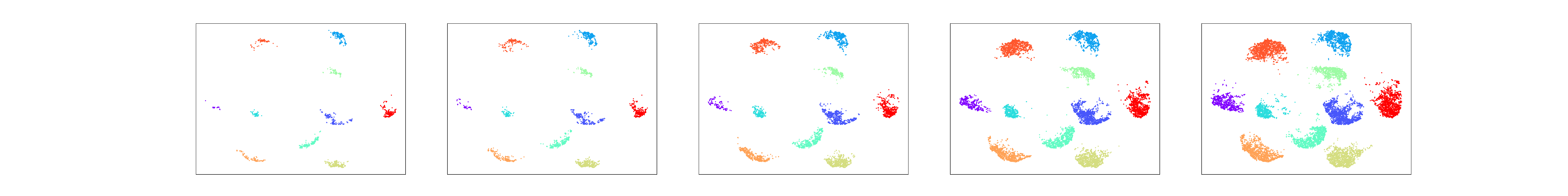}
    }
    \subfigure[]{
        \includegraphics[width=\linewidth]{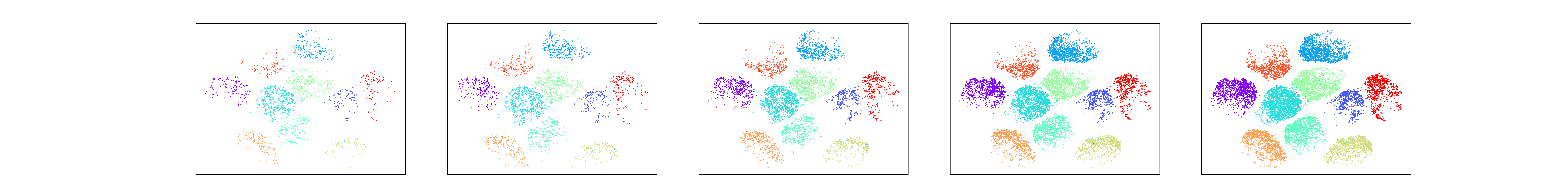}
    }
    \subfigure[]{
        \includegraphics[width=\linewidth]{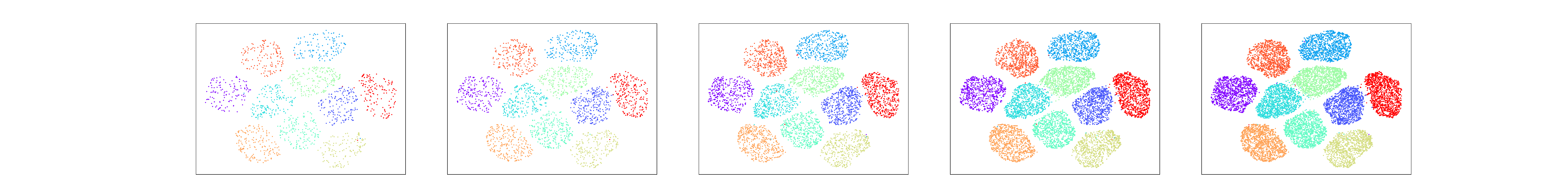}
    }
    \subfigure[]{
        \includegraphics[width=\linewidth]{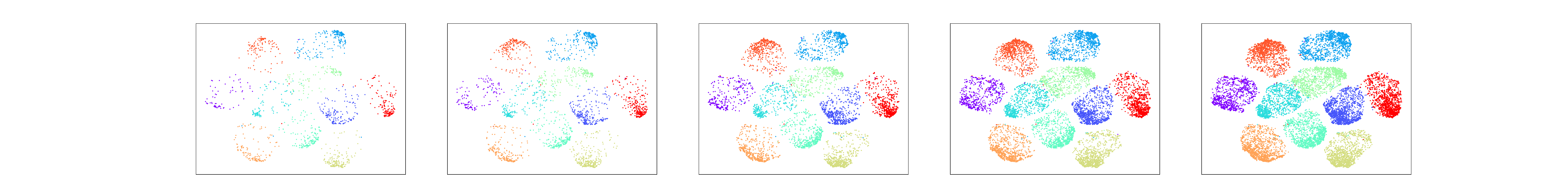}
    }
    \caption{Visualization of predictive distribution of examples through t-SNE: (from left to right column) the number of examples in proxy data, \ie, $k = \{$1.5, 2.5, 5, 10, 15$\}$K; (a) CIFAR-10, (b) Random selection, (c) Entropy-based top-$k$ selection, (d) Entropy-based bottom-$k$ selection, (e) Forget events-based selection, (f) Greedy $k$-center selection, and (g) Proposed selection in this study (i.e., probabilistic selection following $P_1$ defined in the main text). }
    \label{appendix_fig:tsne_merged}
    \vspace{-1.0em}
\end{figure*}

We also attach \figurename~\ref{appendix_fig:tsne_merged} which shows t-SNE~\cite{maaten2008tsne} visualizations of the predictive distributions of examples in all proxy data of CIFAR-10.


\section{Supplemental Results of NAS Benchmarks}
\begin{figure*}[t]
    \centering
    \includegraphics[width=\linewidth]{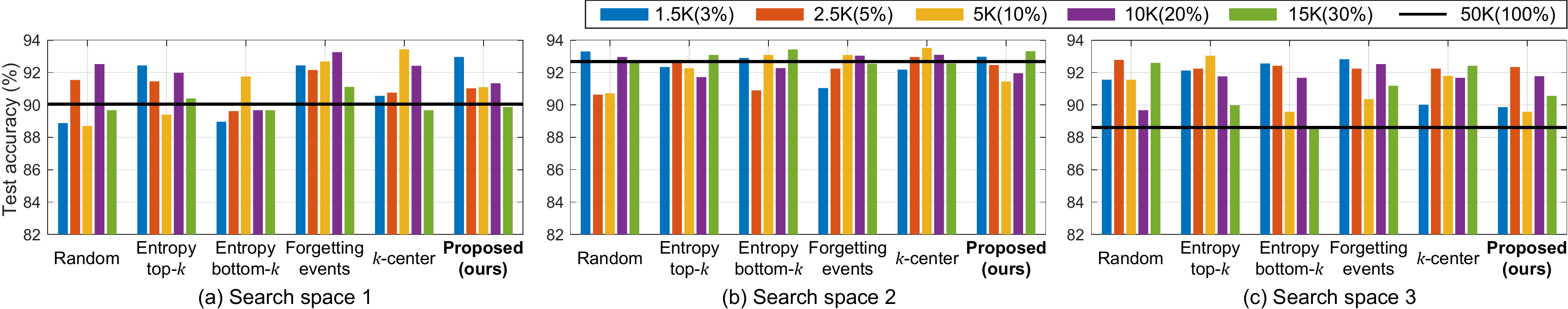}
    \caption{Search performance (CIFAR-10 test accuracy) on NAS-Bench-1shot1 (ENAS) with various proxy data. \textbf{Proposed (ours)} indicates the proposed selection method in Section~\ref{sec:proposed}, specifically referred as a probabilistic selection method following $P_{1}$..}
    \label{appendix_fig:enas}
\end{figure*}
\begin{figure*}[t]
    \centering
    \includegraphics[width=\linewidth]{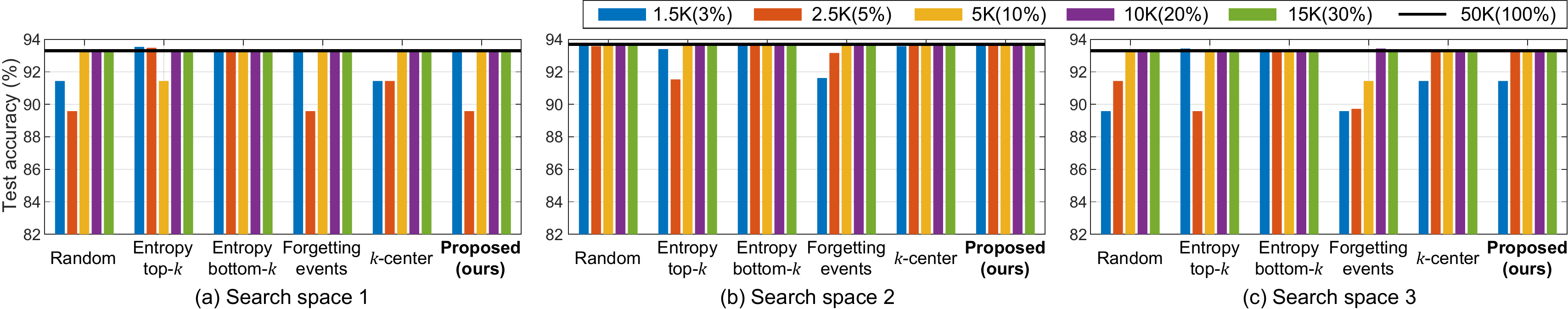}
    \caption{Search performance (CIFAR-10 test accuracy) on NAS-Bench-1shot1 (GDAS) with various proxy data. \textbf{Proposed (ours)} indicates the proposed selection method in Section~\ref{sec:proposed}, specifically referred as a probabilistic selection method following $P_{1}$.}
    \label{appendix_fig:gdas}
\end{figure*}
Using 50K training examples of CIFAR-10, we prepared subsets of varying sizes $\{$1.5, 2.5, 5, 10, 15$\}$K constructed by five different selection functions.
We evaluate DARTS~\cite{yang2019darts}, ENAS~\cite{dean2018enas}, and GDAS~\cite{dong2019searching} on three different search spaces defined by NAS-Bench-1shot1.
Results of DARTS are available in the main paper (\figurename~\ref{fig:nas-bench-1shot1_darts}), and those of the others are following.

\subsection{ENAS}

The results of ENAS are shown in \figurename~\ref{appendix_fig:enas}, and are different from results of DARTS search experiments.
In most subset configurations, the searched architectures using proxy data are better than that using the entire target data, confirming the validity of our proxy data approach.
However, correlation between the search performance and the sizes (or types) proxy data is unclear, and the search performance appears to be less stable compared to that of DARTS.
Given a more stable ENAS-based search algorithm, more reliable observations may be drawn from the experiments.

\subsection{GDAS}

The results of GDAS are shown in \figurename~\ref{appendix_fig:gdas} and confirm again the validity of our proxy data approach in NAS.

\subsection{Discussion on Class Balance}
\begin{figure*}[t]
    \centering
    \includegraphics[width=\linewidth]{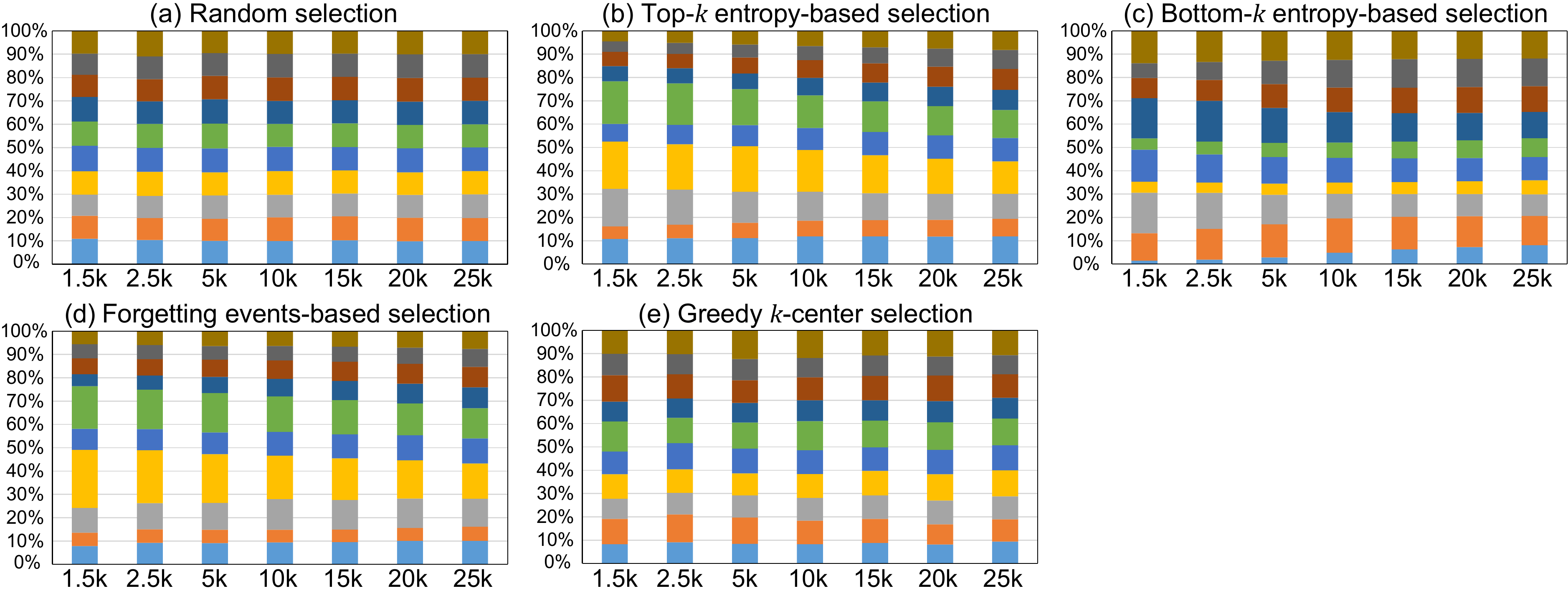}
    \caption{Class ratio of all the proxy datas}
    \label{appendix_fig:class_ratio}
\end{figure*}
\begin{figure*}[t]
    \centering
    \includegraphics[width=\linewidth]{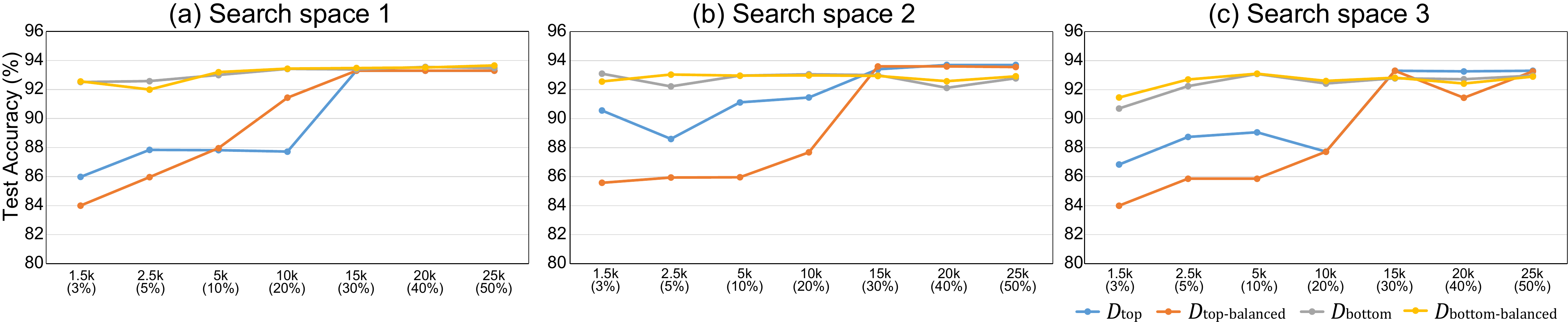}
    \caption{Effects on search performance of the balanced class ratio in entropy-based selection}
    \label{appendix_fig:class_bal}
\end{figure*}

\figurename~\ref{appendix_fig:class_ratio} presents class distributions of all the proxy datarm configurations used in exploration study on NAS-Bench-1shot1.
$D_\mathrm{random}$ and $D_\mathrm{center}$ have class ratios similar to CIFAR-10 (\ie, uniform ratio), and class ratios of the other proxy data are more skewed.

To investigate effects of such class skewedness in search performance, we constructed class-balanced subsets which also satisfy the entropy-based selection rule for top-$k$ and bottom-$k$.
The results on NAS-Bench-1shot1 (DARTS) are visualized in \figurename~\ref{appendix_fig:class_bal}.
In $D_\mathrm{bottom}$, the balanced class ratio has little effect on search performance regardless of sizes.
In contrast, the balanced class ratio in $D_\mathrm{top}$ has negative effect on search performance, especially when $k \leq 10$K.
Therefore, we deduce that entropy distribution in proxy data is more important than the balanced class ratio to maintain search performance.

\subsection{How to Assign Training and Validation data}
\begin{figure*}[t]
    \centering
    \includegraphics[width=\linewidth]{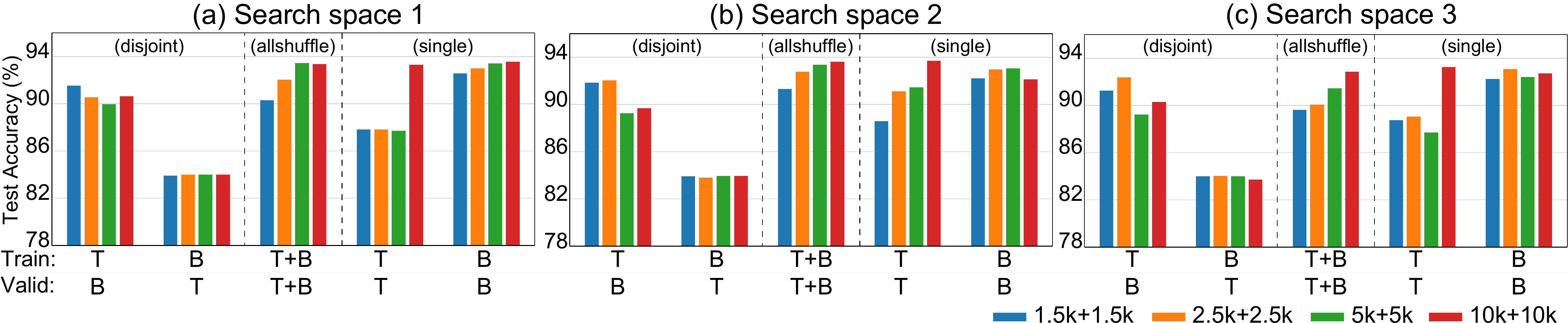}
    \caption{Test accuracies of neural networks searched with combination of two subsets constructed by top-$k$ and bottom-$k$ entropy-based selections. T and B denote $D_\mathrm{top}$ and $D_\mathrm{bottom}$, respectively.}
    \label{fig:hybrid_one_to_one}
\end{figure*}

NAS can be formulated as a bi-level optimization problem~\cite{yang2019darts} as follows:
%
\begin{equation}\label{eq:NAS_formula}
    \begin{array}{l}
    \alpha^{\ast} = \argmin_{\alpha} L_\mathrm{val} (\alpha, w^{\ast}(\alpha) ; D_\mathrm{val}) \\
    \textnormal{s.t.}~ w^{\ast}(\alpha) = \argmin_{w} L_\mathrm{tr} (w ; D_\mathrm{tr}),
    \end{array}
\end{equation}
where $\alpha^\ast$ is the optimal architecture that minimizes a target objective, $w$ is a set of trainable weights in a neural network, and $L_\mathrm{tr}$ and $L_\mathrm{val}$ are the training and validation losses, respectively.
It is typical for NAS algorithms to arbitrarily split the target data into two parts evenly, such that $|D_\mathrm{tr}| = |D_\mathrm{val}|$ in Eq.~\ref{eq:NAS_formula}.
Prior to evaluating the proposed deterministic selection, we do not follow this common practice to find the most suitable split strategy of $D_\mathrm{tr}$ and $D_\mathrm{val}$ within proxy data.
Given $\beta=\frac{1}{2}$, we test two settings of assigning $D_\mathrm{tr}$ and $D_\mathrm{val}$: 1) \textit{disjoint setting}, in which examples with top-$\frac{1}{2}k$ and bottom-$\frac{1}{2}k$ are disjointly assigned to $D_\mathrm{tr}$ and $D_\mathrm{val}$ and 2) \textit{allshuffle setting}, in which examples with top-$\frac{1}{2}k$ and bottom-$\frac{1}{2}k$ are completely shuffled and then split into $D_\mathrm{tr}$ and $D_\mathrm{val}$.

\figurename~\ref{fig:hybrid_one_to_one} compares the results of the disjoint and the allshuffle settings to a \textit{single setting}, which refers to results with proxy data constructed by either entropy top-$k$ or bottom-$k$ selection.
The search performance of the disjoint setting is generally the worst among the three settings.
Especially, on the disjoint setting, search performance is very poor in the cases of assigning examples with top-$\frac{1}{2} k$ to $D_\mathrm{val}$, indicating that using only high-entropy examples for updating architectural parameters leads to significant search performance degradation.
We deduce that the entropy distribution of examples in $D_\mathrm{tr}$ and $D_\mathrm{val}$ must be consistent to achieve higher search performance.
Hence, we adopt the allshuffle settings for the proposed deterministic selection reported in the main paper.

\subsection{DARTS with Proposed Selection Method}
\begin{figure*}[t]
    \centering
    \includegraphics[width=\linewidth]{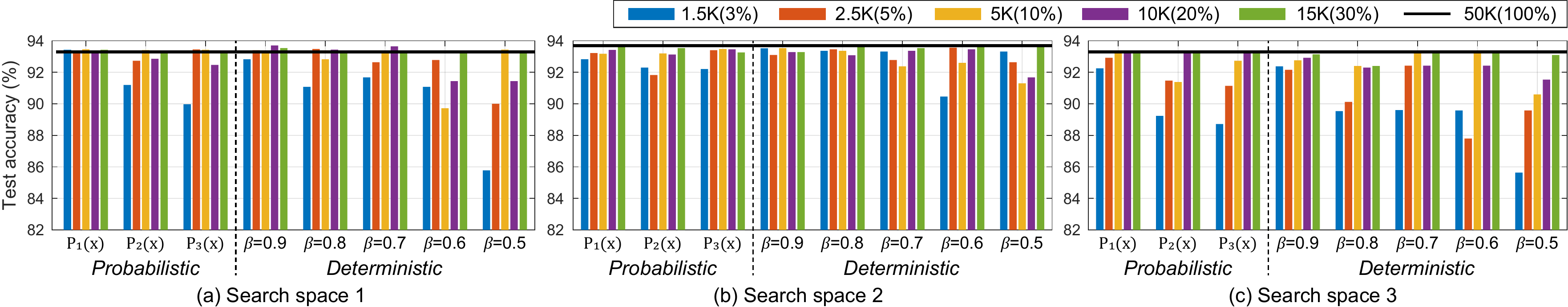}
    \caption{Search performance (CIFAR-10 test accuracy) on NAS-Bench-1shot1 (DARTS) with the proposed selections that are implemented by two manners; probabilistic and deterministic. The probabilistic selection based on $P_1(x)$ is considered as the main method in this study.}
    \label{appendix_fig:probabilistic_vs_deterministic}
\end{figure*}
The proposed selection method is implemented by two approaches in the main text (Section~\ref{sec:proposed}).
For deterministic selection, we conducted experiments with varying $\beta = \{0.9, 0.8, 0.7, 0.6, 0.5\}$.
For probabilistic selection, three selection probabilities defined in the main text were employed.
The results on NAS-Bench-1shots1 (DARTS) are visualized in \figurename~\ref{fig:histogram}.
Among the probabilistic selections, the selection based on $P_1(x)$ achieves the best search performance.
Among the deterministic selections, the selection with $\beta=0.9$ achieves the best search performance.

\subsection{Results of NATS-Bench}
\begin{table}[t]
    \centering
    \caption{Evaluation (top-1 test acc.(\%)) on NATS-Bench using CIFAR-10.}
    \renewcommand{\arraystretch}{1.2}
        \begin{tabular}{c|c|ccc}
            \toprule
            \textbf{NAS} & \textbf{50K} & \textbf{10K} & \textbf{15K} & \textbf{25K} \\
            \midrule
            \textbf{ENAS} & 93.76 & 91.27 & 93.65 & 93.76 \\
            \hline
            \textbf{DARTS} & 54.30 & 54.30 & 54.30 & 54.30 \\
            \bottomrule
        \end{tabular}
    \label{table:nats_bench}
    \vspace{-0.7em}
\end{table}
\tablename~\ref{table:nats_bench} shows the results of NATS-Bench, where we use ENAS and DARTS with the proposed selection.
The search process for each experiment is repeated five times with different seeds, and the average values are reported.
ENAS achieves the original performance when $k \geq 15$K, while DARTS does in all proxy data.

\section{Supplemental Materials for Experiments}
As one of supplementary materials, we have included the code, \ie, code.zip, which is based on DARTS~\cite{yang2019darts}.
Codes for the remaining NAS algorithms, which are included in \tablename~\ref{table:cell_based_real_nas}, will be made public on github after the review process, by extending the our code base.
Instead, in genotypes.py in code.zip, we provide several architectures discovered using NAS algorithms with the proposed selection on CIFAR-10 and ImageNet.
In addition, we share the entropy values of all datasets except ImageNet due to the size restriction policy of IJCAI for supplementary material; it will be also made public after the review process.
For evaluation of the proposed selection method, we utilize four datasets: CIFAR-10, CIFAR-100~\cite{krizhevsky2009cifar}, SVHN~\cite{netzer2011svhn}, and ImageNet~\cite{russakovsky2015imagenet}.
Herein, we review the search spaces and NAS algorithms used in our experiments (Section~\ref{sec:results}).

\subsection{Search Spaces}
\subsubsection{Cell-based Search Space}
A cell-based search space was first introduced by NASNet~\cite{zoph2018nasnet}.
Inspired by the design of GoogLeNet~\cite{szegedy2015going}, which stacks multiple Inception modules repeatedly, the cell-based NAS algorithms search for a single, repeatable cell structure, instead of the entire neural network.
The cell structure, which can be considered as a directed acyclic graph, consists of nodes and edges.
The nodes represent a latent feature map generated by hidden layers, and the edges are associated with operations selected by NAS.
There are two types of cells, \ie, normal and reduction cells.

The cell-based search space was slightly modified~\cite{yang2019darts} by removing weak operations, which hardly are selected by NAS.
The modified search space is utilized generally in recent cell-based NAS algorithms; in this study, the modified search space is referred to as the cell-based search space.
This usually contains eight candidate operations as follows:
\begin{itemize}
    \item zero (a.k.a., none, which is not used when discretizing trained mixed operations in DARTS)
    \item 3x3 separable convolution (x2)
    \item 5x5 separable convolution (x2)
    \item 3x3 dilated separable convolution
    \item 5x5 dilated separable convolution
    \item 3x3 max pooling
    \item 3x3 average pooling
    \item skip connection (a.k.a., identity)
\end{itemize}
In some of NAS algorithms, zero operation was often excluded from the cell-based search space, because it may have little helpful effect on NAS algorithms~\cite{Zela2020rdarts}.

\subsubsection{Restricted Cell-based Search Spaces (S1, S2, S3, and S4)}
To analyze the robustness of DARTS, these four search spaces were introduced ~\cite{Zela2020rdarts}.
They differ from the cell-based search space in the number and types of candidate operations associated on edge inside the cell structure; the super-network structure remains as used in differentiable NAS algorithms.
Each search space is considered as a subset of the cell-based search space, and described as below.
\begin{itemize}
    \item \textbf{S1:} Candidate operations are top-2 operations identified using DARTS search process, and thus differs in all edges.
    \item \textbf{S2:} Candidate operations are $\{$3x3 separable convolution (x2), skip connection$\}$.
    \item \textbf{S3:} Candidate operations are $\{$3x3 separable convolution (x2), skip connection, zero$\}$.
    \item \textbf{S4:} Candidate operations are $\{$3x3 separable convolution (x2), noise$\}$. The noise operation generates a random Gaussian noise $\mathcal{N}(0,1)$ whose size is identical to the input.
\end{itemize}

\subsection{Used NAS Algorithms for Evaluation}

\subsubsection{DARTS}
DARTS~\cite{yang2019darts} introduced the differentiable NAS framework that is usually utilized as a baseline in the differentiable NAS algorithms. 
By employing continuous architecture parameters to the super-network, DARTS solves bi-level optimization (Eq~\ref{eq:NAS_formula}) to optimize the architecture parameters.
In the inner update of the bi-level optimization, weight parameters of operations in the super-network are trained, by using the half of data, namely training data in NAS.
In the outer update, architecture parameters are trained, by using the other half of data, namely validation data in NAS.
After search process that is training process of the super-network according to the bi-level optimization, DARTS selects top-1 operation for every edge in the cell, meaning that the selected operation is associated by the maximum architecture parameter in each edge.

\subsubsection{PC-DARTS}
PC-DARTS~\cite{xu2020pcdarts} proposed partial channel connection to reduce the memory cost used in DARTS search process, and edge normalization to produce more consistent search results in DARTS.
On CIFAR-10, only 1/4 features (outputs of 1/4 convolution filters) in the output of the edges were selected to solve the bi-level optimization.
It resulted in that the batch size of 64 in DARTS increases to 256, and thus accelerated the search process.
Like the evaluation in this study, PC-DARTS searched for architectures on ImageNet with 12.5\% of examples randomly sampled from the ImageNet.
It took 3.8 GPU days, \ie, 11.5 hours with eight V100 GPUs, with batch size of 1024; we speculate that the parallel execution on the eight GPUs would have caused non-negligible overhead.

\subsubsection{EcoNAS (EcoDARTS)}
EcoNAS~\cite{zhou2020econas} suggested four reduction factors: the number of channels in the super-network, the resolution of input images, the number of training epochs during search process, and the sample ratio of the target data.
By investigating the effects of the reduction factors, three of them except the sample ratio were important to obtain rank correlation between the performance of the target network and that of the smaller network.
Based on these findings, EcoNAS proposed a hierarchical search strategy to improve the accuracy of architecture performance estimation using the smaller super-network.
EcoNAS briefly reported the search result using a subset randomly sampled from CIFAR-10, therefore, we extensively evaluate the various sample ratios and selection methods on NAS algorithms.
EcoDARTS(-c4r2) used in this study is a variant of DARTS, where the super-network is reduced by reducing the number of channels from 36 to 8 and the resolution of the input images from 32 to 16.

\subsubsection{SDARTS}
SDARTS~\cite{chen2020stabilizing-darts} proposed a method to perturb architecture parameters during DART search process, in order to overcome instability of DARTS.
Two types of perturbation were suggested: random smoothing (RS) and adversarial training-based perturbation (ADV).
In this study, we utilized SDARTS with RS which is simple yet effective to improve the search performance of DARTS.
In the search process with RS, the architecture parameters are perturbed by adding random Gaussian noise.

\subsubsection{SGAS}
SGAS~\cite{li2020sgas} employed an efficient search strategy which is progressively shirking the search space during DARTS search process.
After training weight parameters of the super-network alone for 10 epochs, SGAS chooses one edge every five epochs according to their proposed selection criteria measuring edge importance.
The chosen edge is discretized, meaning that all candidate operations except the best operation in this edge are removed, and then the search process continues with the remaining non-discretized edges until the all edges in the cell are discretized.

\subsubsection{RobustDARTS}
By introducing the restricted cell-based search spaces (S1-S4), RobustDARTS~\cite{Zela2020rdarts} demonstrated that DARTS failures usually rise due to overfitting of the architecture parameters to the validation data.
To remedy this issue, RobustDARTS regularized the inner objective of the bi-level optimization, by using data augmentation (cutout, scheduled-DropPath) or increasing weight decay factors for L2 regularization.
In this study, we utilized the RobustDARTS with L2 regularization where weight decay factor is 0.0243; the factor is 0.0003 in DARTS.

\subsection{Experimental Settings (Except ImageNet)}

\subsubsection{Search}
Using a single GeForce RTX 2080ti GPU, we execute search process of differentiable NAS algorithms following protocols reported in each algorithm.
These protocols are commonly based on DARTS~\cite{yang2019darts}.
First, we evenly split the proxy data into two parts: one for updating network parameters ($w$) and the other for updating architecture parameters ($\alpha$).

Following usual settings of differentiable NAS algoriths, the super-network is constructed by stacking eight cells (six normal cells and two reduction cells) including mixed operations proposed by DARTS, after 3x3 convolution-based stem layers with 16 initial channels.
Search process is executed for 50 epochs, with a batch size of 64 (exceptionally, 224 in PC-DARTS).
Weight parameters of the super-network are optimized by momentum SGD, with an initial learning rate of 0.1, a momentum of 0.9, and a weight decay of 0.0003, where the learning rate is annealed down to zero with a cosine schedule.
Architecture parameters $\alpha$ are optimized by the Adam optimizer, with a fixed learning rate of 0.0006, a momentum of (0.5, 0.999), and a weight decay of 0.001.
During search process, the standard data preprocessing and augmentation techniques were used: the channel normalization, the central padding of images to 40×40 and then random cropping back to 32×32, random horizontal flipping. 

\subsubsection{Final evaluation}
We follow DARTS settings, and a single GeForce RTX 2080ti GPU is used to train the discovered neural architectures.
The network is constructed by stacking 20 cells (18 normal cells and two reduction cells, each type of which share the same network discovered in search process) after 3x3 convolution-based stem layers with 36 initial channels, including an auxiliary loss.
In training process, the network is trained from scratch for 600 epochs using training examples, with a batch size of 96; several architectures, which cannot use the batch size of 96 due to memory size, are trained with a batch size of 84.
The momentum SGD optimizer is used, with an initial learning rate of 0.025 following cosine scheduled annealing, a momentum of 0.9, a weight decay of 0.0003, and a norm gradient clipping at 5.
With aforementioned data augmentation techniques, cutout is additionally used, and drop-path with a rate of 0.2 is used for regularization.

\subsubsection{Final evaluation on cells searched on S1-4}
Following RobustDARTS settings~\cite{Zela2020rdarts}, this evaluation keeps all the settings for final evaluation, except network configurations.
For CIFAR-10 on S1 and S3, the networks stacks 20 cells with initial channels of 36.
For CIFAR-10 on S2 and S4, the networks stacks 20 cells with initial channels of 16.
For CIFAR-100 and SVHN on S1-4, the networks stacks 8 cells with initial channels of 16.

\subsection{Experimental Settings on ImageNet}

\subsubsection{Search}
We execute the search process on a single Tesla V100 GPU.
The super-network is nearly identical to the aforementioned super-network used in CIFAR-10, CIFAR-100, and SVHN, with slight modifications.
To fit ImageNet search on GPU memory, stem layers in the proxy model consists of three 3x3 convolution layers of stride 2, resulting in that the input data resolution of the first cell is reduced from 224$\times$224 to 28$\times$28.
In addition, the batch size is set as 256 on both DARTS and PC-DARTS with the proposed proxy data selection.
All of the other hyperparamters are not changed.

Note that PC-DARTS, which directly executed searching process on ImageNet as in this study, used eight V100 GPUs and the batch size is 1,024.
The authors of PC-DARTS used $D_\mathrm{tr}$ and $D_\mathrm{val}$ in Eq.~\ref{eq:NAS_formula} which are randomly sampled from ImageNet training set with 10\% and 2.5\% examples, respectively.

\subsubsection{Final evaluation}
We follow PC-DARTS~\cite{xu2020pcdarts} settings, but we only used two V100 GPUs for ImageNet training with batch size of 640.
The network is constructed by stacking 14 cells (12 normal cells and two reduction cells, each type of which share the same network discovered in search process) after the stem layers which are three 3x3 convolution layers of stride 2, with 48 initial channels and an auxiliary loss.
In training process, the network is trained from scratch for 250 epochs using training 1.28M examples of ImageNet.
The momentum SGD optimizer is used, with an initial learning rate of 0.5 which is decayed down to zero linearly, a momentum of 0.9, a weight decay of 0.00003, and a norm gradient clipping at 5.
During the first five epochs, the learning rate warm-up is applied.
Label smoothing with a rate of 0.1 is used to enhance the training.

\subsection{Comparison with Existing Selection Methods}
\begin{table}[t]
    \centering
    \renewcommand{\arraystretch}{1.1}
        \begin{tabular}{c|cc}
            \toprule
            \textbf{Selection} & \textbf{Proxy data size} & \textbf{Test Err. (\%)} \\
            \midrule
            Random & 10K & 2.95  \\
            Entropy top-$k$ & 10K & 3.04  \\
            Entropy bottom-$k$ & 10K & 2.97  \\
            Forget events & 10K & 3.14  \\
            $k$-center & 10K & 3.36  \\
            \midrule
            Proposed & 10K & 2.92 \\
            \bottomrule
        \end{tabular}
    \caption{Evaluation (top-1 test error (\%)) on DARTS with existing selection methods and CIFAR-10.}
    \label{table:selections_darts}
    \vspace{-0.7em}
\end{table}

In Section~\ref{sec:exploration}, we explored search performance with existing selection methods on NAS-Bench-1shot1.
Likewise, we evalute the existing selection methods on the cell-based search space using DARTS.
The search results are shown in \tablename~\ref{table:selections_darts}, and the proposed selection achieves the higher search performance than the others.

\subsection{Comparison with NAS Algorithms on CIFAR-10}
\begin{figure}
    \centering
    \includegraphics[width=\linewidth]{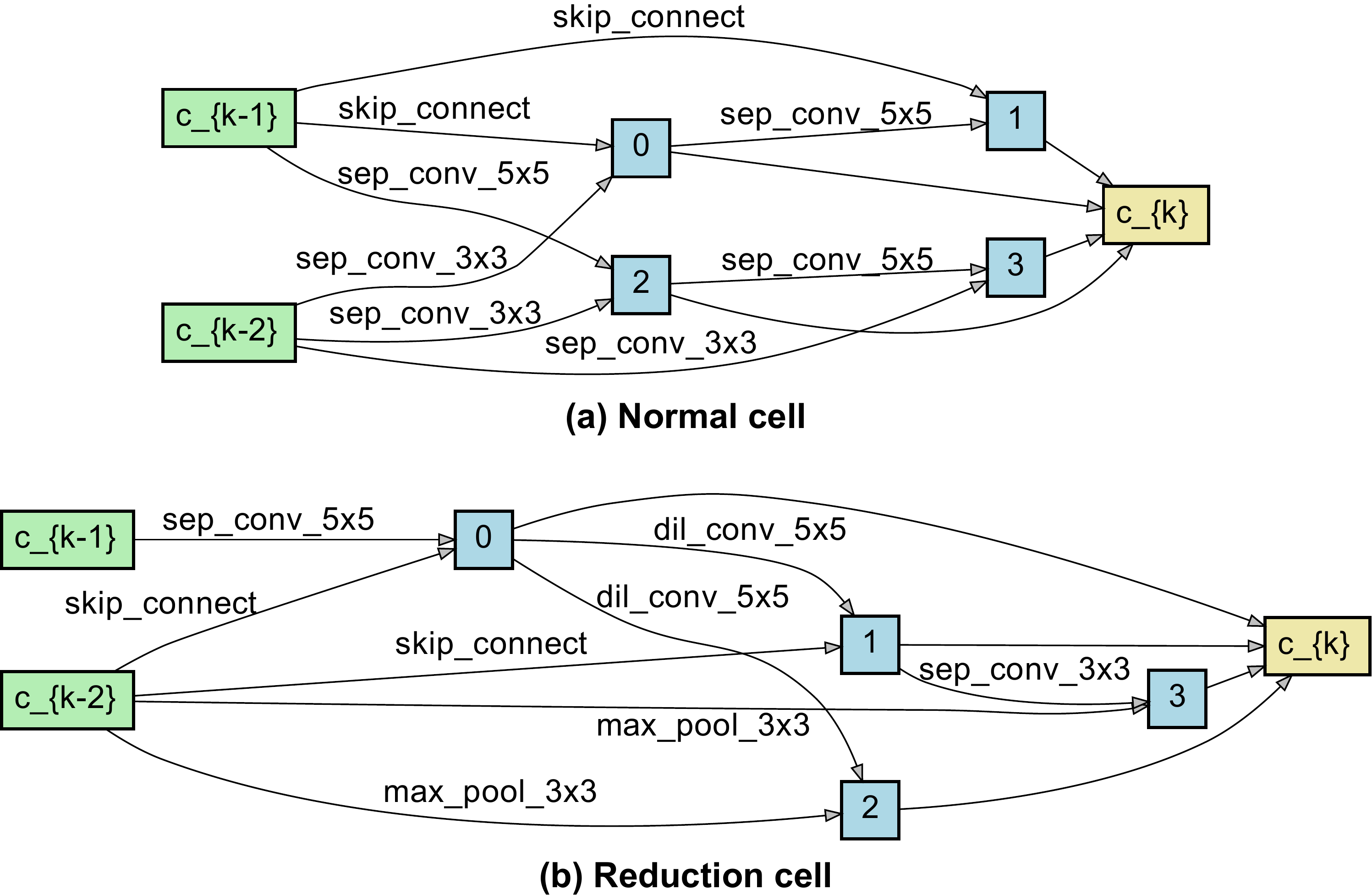}
    \caption{The cell structures discovered using DARTS with the proposed selection (proxy data, 10\% of ImageNet).}
    \label{fig:imagenet_cells}
\end{figure}
In the main text, we successfully searched for the architecture on ImageNet with the proposed selection; the cell structures are shown in \figurename~\ref{fig:imagenet_cells}
Notably, this search cost is similar to the search cost of other NAS algorithms using entire CIFAR-10, as shown in the column of search cost in \tablename~\ref{table:CIFAR-10}.
Under the fair search cost, we can evaluate our discovered architecture by transferring from ImageNet to CIFAR-10.
As shown in \tablename~\ref{table:CIFAR-10}, this architecture yields top-1 test error of \textbf{2.4\%} on CIFAR-10 when the batch size is 84, which is the state-of-the-art performance among cell-based NAS algorithms.
\input{tex/8_appendix_cifar10}

%% file: tex/8_appendix_cifar10.tex
\begin{table*}[h]
    \begin{threeparttable}
    \centering
    \caption{Comparison with state-of-the-art neural architectures on CIFAR-10. In search method, RL and EA indicate NAS algorithms based on reinforcement learning and evolutionary algorithm, respectively. The gradient-based NAS algorithms means the differentiable NAS algorithms in this study.}
    \renewcommand{\arraystretch}{1.1}
        \begin{tabular}{lcccc}
            \toprule
            \textbf{Neural Architectures} & \textbf{Top-1 Test Err. (\%)} & \textbf{Params. (M)} & \textbf{Cost (GPU days)} & \textbf{Search Method}\\
            \hline
            DenseNet-BC$^\ast$~\cite{huang2017densely} & 3.46 & 25.6 & - & manual \\
            \hline
            NASNet-A~\cite{zoph2018nasnet} & 2.65 & 3.3 & 1800 & RL \\
            AmoebaNet-A~\cite{real2019regularized} & 3.34 $\pm$ 0.06 & 3.2 & 3150 & EA \\
            ENAS~\cite{dean2018enas} & 2.89 & 4.6 & 0.5 & RL \\
            EcoNAS~\cite{zhou2020econas} & 2.62 $\pm$ 0.02 & 2.9 & 8 & EA \\
            \hline
            DARTS (1st order)~\cite{yang2019darts} & 3.00 $\pm$ 0.14 & 3.2 & 0.26$^\ddag$ & gradient \\
            DARTS (2nd order)~\cite{yang2019darts} & 2.76 $\pm$ 0.09 & 3.3 & 1.06$^\ddag$ & gradient  \\
            SNAS (moderate)~\cite{liang2019snas} & 2.85 $\pm$ 0.02 & 2.8 & 1.5 & gradient \\
            GDAS~\cite{dong2019searching} & 2.93 & 3.4 & 0.21 & gradient \\
            PDARTS~\cite{chen2019pdarts} & 2.5 & 3.4 & 0.3 & gradient \\
            PC-DARTS~\cite{xu2020pcdarts} & 2.67 $\pm$ 0.07 & 3.6 & 0.08$^\ddag$ & gradient \\
            EcoDARTS-c4r2~\cite{zhou2020econas} & 2.80 & 4.5 & 0.23$^\ddag$ & gradient \\
            SGAS-Cri.1~\cite{li2020sgas} & 2.66 $\pm$ 0.24 & 3.7 & 0.19$^\ddag$ & gradient \\
            SDARTS-RS~\cite{chen2020stabilizing-darts} & 2.67 $\pm$ 0.03 & 3.4 & 0.23$^\ddag$ & gradient \\
            FairDARTS~\cite{chu2020fairdarts} & 2.54 $\pm$ 0.05 & 2.8 & 0.4 & gradient \\
            \hline
            Proposed selection (10\%, CIFAR-10) & & & & \\
            ~~ + PC-DARTS      & 2.91 $\pm$ 0.03 & 3.4 $\pm$ 0.3 & 0.01$^\ddag$ & gradient \\
            ~~ + EcoDARTS-c4r2 & 2.81 $\pm$ 0.07 & 3.6 $\pm$ 0.2 & 0.02$^\ddag$ & gradient \\
            ~~ + SDARTS-RS     & 2.83 $\pm$ 0.15 & 3.3 $\pm$ 0.2 & 0.03$^\ddag$ & gradient \\
            ~~ + SGAS-Cri.1    & 2.72 $\pm$ 0.08 & 3.7 $\pm$ 0.2 & 0.02$^\ddag$ & gradient \\
            \hline
            DARTS (1st) & & & & gradient \\
            ~~ + Proposed selection (10\%, CIFAR-10) & 2.94 $\pm$ 0.21 & 3.7 $\pm$ 0.3 & 0.03$^\ddag$ & \\ 
            ~~ + Proposed selection (20\%, CIFAR-10) & 2.92 $\pm$ 0.30 & 3.5 $\pm$ 0.4 & 0.05$^\ddag$ & \\ 
            ~~ + Proposed selection (30\%, CIFAR-10) & 2.88 $\pm$ 0.01 & 3.0 $\pm$ 0.2 & 0.07$^\ddag$ & \\ 
            ~~ + Proposed selection (40\%, CIFAR-10) & 2.78 $\pm$ 0.16 & 3.2 $\pm$ 0.2 & 0.10$^\ddag$ & \\ 
            ~~ + Proposed selection (50\%, CIFAR-10) & 2.76 $\pm$ 0.14 & 3.2 $\pm$ 0.3 & 0.13$^\ddag$ & \\ 
            ~~ + \textbf{Proposed selection (10\%, ImageNet)} & \textbf{2.4} & \textbf{3.8} & \textbf{0.26}$^\dagger$ \\
            \bottomrule
        \end{tabular}
        \begin{tablenotes}
            \small
            \item $^\ddag$denotes the search cost with a single GeForce RTX 2080ti GPU
            \item $^\dagger$denotes the search cost with a single Tesla V100 GPU when DARTS is executed on ImageNet with the proposed selection (where 10\% of examples in ImageNet are selected). The others (not denoted by $^\ddag$ or $^\dagger$) are the search costs, which reported in original papers, when the algorithms were executed on CIFAR-10.
            \item $^\ast$denotes neural networks trained without cutout; the others used cutout for additional data augmentation.
        \end{tablenotes}
    \label{table:CIFAR-10}
    \end{threeparttable}
\end{table*}